\documentclass[perprint
,10pt]{elsarticle}

\usepackage{amsmath,amssymb,amsfonts}
\usepackage{graphicx}
\usepackage{xcolor}
\usepackage{booktabs}
\usepackage{hyperref}
\usepackage{algorithm}
\usepackage{algpseudocode}
\usepackage{multirow}

\journal{Computers \& Education: Artificial Intelligence}

\begin{document}

\begin{frontmatter}

\title{MC-CPO: Mastery-Conditioned Constrained Policy Optimization for Pedagogically Safe Intelligent Tutoring Systems}

\author[1]{Oluseyi Olukola}
\author[1]{Nick Rahimi}
\affiliation[1]{School of Computing Sciences and Computer Engineering, 
University of Southern Mississippi, Hattiesburg, MS 39406, USA}

\begin{abstract}
Intelligent tutoring systems increasingly rely on reinforcement learning to
personalise instruction, yet optimising for observable engagement signals task completion, correctness, session length can systematically decouple
learner activity from genuine knowledge acquisition. Analysing over 21~million
student interactions across two large-scale deployed platforms, we find that
engagement events without corresponding mastery gains occur in 26.5\% of
interactions on Junyi Academy (72,758 students, 16.2M interactions) and 3.1\%
on XES3G5M (14,453 students, NeurIPS~2023 benchmark), confirming that this
engagement--learning decoupling is directly observable in real educational
technology at scale, and not a simulation artefact.
 
We introduce \textbf{Mastery-Conditioned Constrained Policy Optimisation
(MC-CPO)}, a reinforcement learning framework that addresses this problem
structurally rather than through reward engineering. MC-CPO conditions the
admissible instructional action space directly on learner mastery state: a
concept becomes available only when all prerequisite knowledge meets a
mastery threshold, yielding an action space that expands naturally as learners
acquire knowledge. Pedagogical safety constraints are enforced by construction,
with formal guarantees of structural prerequisite safety, primal--dual convergence,
and strict dominance over post-hoc filtering.
 
Empirical evaluation across simulated environments and both real-world
datasets demonstrates that MC-CPO is the only method to reduce reward hacking
severity across all conditions. Mean per-episode mastery gain (\(\Delta K\))
increases by 18.3\% on Junyi Academy and 54.0\% on XES3G5M relative to
all baselines, while competitive engagement performance is maintained. Reward
shaping and post-hoc filtering fail to separate from unconstrained baselines
in all settings. These results indicate that structural constraint modelling embedding pedagogical feasibility directly into the policy space provides a
principled and empirically validated foundation for safer adaptive instructional
policies in deployed intelligent tutoring systems.
 
\end{abstract}


\end{frontmatter}


\section{Introduction}
\label{intro}

Intelligent tutoring systems (ITS) have demonstrated substantial effectiveness at supporting learner outcomes across a wide range of educational contexts \cite{VanLehn2011ITS,Koedinger2006CognitiveTutors}. A central design challenge in these systems is the specification of an instructional policy: a principled mapping from learner state to the next concept or activity to present. As ITS platforms scale, reinforcement learning (RL) has increasingly been explored \cite{SuttonBarto2018} as a mechanism for adaptive instructional sequencing, personalization, and policy optimization. In principle, RL provides a natural framework for modeling long-horizon decision-making in educational environments, where instructional actions influence future learner states and outcomes.

However, a fundamental tension arises in practice. Deployable ITS platforms typically optimize for observable engagement signals such as task completion, correctness, or session length because genuine knowledge acquisition is difficult to measure in real time. When a reinforcement learning policy is trained on these proxy signals, it may learn instructional strategies that sustain high engagement without promoting meaningful learning \cite{Doroudi2019WheresReward,Mon2023RLInEducation}. We refer to this phenomenon as \emph{engagement--learning decoupling}: the systematic divergence between what an optimized policy rewards and what it actually teaches.

To establish the scope of this problem in real deployed systems, we analyze two large-scale publicly available ITS datasets. On Junyi Academy a deployed Taiwanese e-learning platform with 72{,}758 students and 16.2 million interactions we find that 26.5\% of student interactions constitute engagement events without a corresponding mastery gain. On XES3G5M \cite{Liu2023XES3G5M}, a NeurIPS 2023 knowledge tracing benchmark with 14{,}453 students and 5.1 million interactions from a Chinese mathematics tutoring platform, the rate is 3.1\%. These rates confirm that engagement learning decoupling is not a simulation artifact but a directly observable pattern in deployed educational technology at scale.

Addressing this problem through reward engineering is difficult. Specifying reward functions that faithfully represent pedagogical goals remains a persistent and non-trivial challenge, and prior work on instructional RL notes that proxy objectives can lead to undesirable instructional policies \cite{Doroudi2019WheresReward}. The reinforcement learning literature has more broadly identified reward hacking as a systematic failure mode arising when agents exploit imperfections in proxy objectives \cite{Amodei2016ConcreteProblems,Skalse2022RewardHacking}. Approaches based on reward modeling, inverse reward design, or empirical detection strategies \cite{HadfieldMenell2017IRD,Shihab2025ProxyGaming} attempt to improve or correct the reward signal rather than structurally constraining the policy space itself.

Concurrently, safe reinforcement learning has developed principled methods for enforcing long-term constraints within the Constrained Markov Decision Process (CMDP) framework \cite{Altman1999CMDP,Puterman1994MDP,Achiam2017CPO}. Modern constrained policy optimization techniques integrate dual variables or Lyapunov-based formulations to balance return maximization with constraint satisfaction \cite{Chow2018Lyapunov,Tessler2019RCPO}. Comprehensive surveys emphasize the diversity of constraint formulations and the importance of state-aware safety mechanisms in dynamic environments \cite{Wachi2024SafeRLSurvey,Zhao2023StatewiseSafeRL}. More recent work explores monotonic expansion of feasible regions as policies improve constraint adherence \cite{Yang2023FPI}. While these contributions provide strong foundations for constraint-aware optimization, constraints in these settings are typically exogenous and static, such as safety thresholds in robotics or resource budgets in control systems. Feasibility expansion in prior work is generally policy-driven, reflecting improved compliance with externally specified constraints.

Educational environments present a structurally distinct challenge. In instructional contexts, what constitutes a feasible instructional action may depend directly on the learner’s evolving mastery state. Prerequisite satisfaction and knowledge acquisition dynamically alter which concepts or tasks are pedagogically appropriate at a given time. In such settings, feasibility is not merely a function of external safety thresholds but can be endogenous to the learner’s internal knowledge trajectory. Despite extensive work on knowledge tracing \cite{CorbettAnderson1994BKT,Piech2015DKT} and graph-based instructional modeling \cite{Deng2023GITS}, to our knowledge there has been limited formal treatment of mastery-conditioned feasibility within a CMDP framework, particularly in relation to reward hacking prevention.

In this paper, we introduce \textbf{MC-CPO: Mastery-Conditioned Constrained Policy Optimization}, a constrained reinforcement learning algorithm in which the feasible action set is explicitly conditioned on the learner’s evolving mastery state. Rather than encoding pedagogical correctness in the reward function, MC-CPO enforces structural feasibility through a mastery-conditioned action space that expands monotonically as prerequisite knowledge is acquired. This design induces a dynamic policy space in which instructional actions become admissible only when pedagogically appropriate, thereby reducing reliance on reward shaping to prevent undesirable behaviors. Importantly, feasibility expansion in MC-CPO is state-driven via mastery progression, rather than policy-driven via constraint satisfaction improvement.

We further formalize pedagogical safety constraints within a CMDP framework and show that mastery-conditioned feasibility induces a state-driven expansion of admissible policies. We prove that MC-CPO enforces structural prerequisite safety by construction and provide a safety gap theorem demonstrating that post-hoc filtering of unconstrained policies can be strictly suboptimal relative to optimization within the feasible set. Finally, we empirically evaluate MC-CPO in a simulated tutoring environment and measure reward hacking severity using a previously defined Reward Hacking Severity Index (RHSI), enabling direct comparison to unconstrained and post-hoc baseline methods.

The primary contributions of this work are as follows:

\begin{itemize}
    \item We introduce MC-CPO, a constrained RL formulation for educational settings in which feasibility is conditioned on learner mastery state, yielding a monotonically expanding policy space that differs from static or policy-driven feasibility in prior safe RL methods.
    
    \item We provide formal guarantees that MC-CPO enforces prerequisite safety structurally and prove a safety gap theorem showing that post-hoc filtering of unconstrained policies cannot achieve the same reward--safety tradeoff as constrained optimization within the mastery-conditioned feasible set.
    
    \item We demonstrate empirically, across a simulated intelligent tutoring environment (15 and 25 concepts) and two real-world ITS datasets Junyi Academy (16.2M interactions, 72{,}758 students) and XES3G5M (NeurIPS 2023, 5.1M interactions, 14{,}453 students) that MC-CPO substantially reduces reward hacking severity (as measured by RHSI) while maintaining competitive engagement performance relative to standard baselines. Real-world analysis confirms that the engagement--learning decoupling targeted by MC-CPO is directly observable in deployed ITS data, occurring at rates of 26.5\% and 3.1\% of student interactions on the two platforms.
\end{itemize}

The paper is organized as follows: Section~\ref{review} reviews related work; Section~\ref{sec:problem_setup} formalizes the CMDP; Section~\ref{sec:algorithm} presents MC-CPO; Section~\ref{sec:theory} develops theoretical guarantees; Section~\ref{sec:experiments} reports empirical evaluations including real-world validation; Sections~\ref{sec:discussion}--\ref{sec:conclusion} discuss and conclude.


\section{Related Work}
\label{review}

Research on safe reinforcement learning (SafeRL) has advanced considerably,
with much of the modern literature adopting the Constrained Markov Decision
Process (CMDP) framework to formalize tradeoffs between return maximization
and constraint satisfaction. Foundational surveys and syntheses tend to
emphasize both the breadth of CMDP formulations and the practical prevalence
of primal-dual and Lagrangian approaches in contemporary safe RL pipelines
\cite{Kushwaha2025Survey,Garcia2015SafeRLSurvey,Wachi2024SafeRLSurvey,Gu2023SafeRLReview}.
More recent overviews further highlight distinctions between static,
state-dependent, and trajectory-level constraints, suggesting that feasibility
definitions can vary substantially across domains \cite{Zhao2023StatewiseSafeRL}.
Within this broader direction, several works explore constraint handling under
limited exploration, including safe exploration methods that project actions
onto feasible sets or otherwise modify policy updates to avoid unsafe regions
\cite{Dalal2018SafeExploration,Wachi2023SafeExploration,Shihab2025ProxyGaming}.

A parallel line of work investigates how feasible regions are identified and
expanded under safety requirements. Feasible Policy Iteration (FPI) proposes
a region-wise improvement principle with theoretical guarantees that the
feasible region expands monotonically as the policy improves its satisfaction
of static safety constraints \cite{Yang2023FPI}. Related developments in this line confirm the practical importance of monotonic feasibility expansion for safe policy learning. Although these approaches appear closely related in spirit
to feasibility expansion, they typically treat constraints as exogenous and
fixed; expansion reflects improved policy compliance with predefined safety
sets rather than endogenous changes in the environment itself. In educational
settings, by contrast, feasibility may evolve as a function of learner mastery,
suggesting a structurally distinct form of state-driven expansion that these
frameworks do not appear to capture.

Work on constraint-conditioned or threshold-adaptive safe RL further
illustrates the diversity of CMDP objectives. Conditioned Constrained Policy
Optimization (CCPO), for example, studies zero-shot adaptation to varying
constraint thresholds at deployment \cite{Yao2023CCPO}. While conceptually
related in its use of constraint conditioning, this setting differs from
mastery-conditioned feasibility in that constraint thresholds remain externally
specified rather than derived from learner state. Additional safe exploration
methods aim to reduce violations during learning, including doubly
optimistic--pessimistic exploration schemes \cite{Bura2021DOPE},
Lyapunov-based safety shaping \cite{Chow2018Lyapunov}, and primal-dual
penalty methods such as Reward Constrained Policy Optimization (RCPO)
\cite{Tessler2019RCPO}. Constrained Policy Optimization (CPO) remains a
widely-used baseline for modern policy-gradient-based constrained optimization
\cite{Achiam2017CPO}, and Proximal Policy Optimization (PPO) serves as a
practical backbone in both unconstrained and constrained variants
\cite{Schulman2017PPO}. Collectively, these works demonstrate the maturity
of constraint-based optimization machinery, while also suggesting that
constraint structure itself may meaningfully shape policy learning dynamics.

A central motivation for pedagogical safety is the broader problem of reward
misspecification and reward hacking. Early AI safety work identifies reward
hacking as a concrete failure mode arising when proxy objectives are optimized
in place of intended outcomes \cite{Amodei2016ConcreteProblems}. More recent
theoretical analyses attempt to formalize reward hacking in reinforcement
learning and characterize conditions under which proxy optimization may diverge
from designer intent \cite{Skalse2022RewardHacking,Pan2022RewardMisspec}.
Related work in inverse reward design and reward model overoptimization further
highlights how learned or proxy rewards may be exploited in ways that were not
anticipated \cite{HadfieldMenell2017IRD,Gao2023ScalingReward}. Empirical
mitigation strategies have also been proposed across diverse RL settings,
including approaches that detect or penalize proxy gaming behaviors
\cite{Shihab2025ProxyGaming,Opryshko2025MCVL}. The potential of reward
shaping as a partial remedy has also been explored, though policy invariance
results suggest that shaping alone may not eliminate structural incentives for
exploitation \cite{Ng1999RewardShaping}. At the same time, developments in
reinforcement learning from human feedback (RLHF) underscore the broader
challenge of aligning reward signals with intended objectives, particularly
when optimization pressures can induce overfitting to imperfect reward models
\cite{Christiano2017RLHF,Ibarz2018RewardModelHacking,Dai2024SafeRLHF}. Taken
together, these strands suggest that improving reward design alone may be
insufficient to fully eliminate structural incentives for exploitation.

In educational reinforcement learning and intelligent tutoring systems (ITS),
RL has been explored for instructional sequencing and personalization, while
also raising concerns about specifying reward signals that faithfully represent
learning goals. The effectiveness of well-designed tutoring systems at
supporting learning has been documented broadly \cite{VanLehn2011ITS,
Koedinger2006CognitiveTutors}, and early POMDP-based formulations suggest that
planning under uncertainty over student knowledge can yield meaningful
instructional improvements \cite{Rafferty2011POMDP}. Reviews of RL in
education, however, note that reward definitions tend to rely on observable
performance or engagement proxies and may not directly encode long-term
knowledge construction \cite{Mon2023RLInEducation,Doroudi2019WheresReward}.
More recent work introduces goal-oriented ITS tasks and graph-based RL
formulations that incorporate prerequisite or concept structures \cite{Deng2023GITS},
yet typically without CMDP-based pedagogical safety constraints or formal
guarantees against reward exploitation. GITS is not included in the empirical comparison as its graph-based reward formulation and partially observable student 
model are not directly compatible with the CMDP experimental protocol used here; a unified benchmark remains an open challenge for the educational RL community. Separately, knowledge-graph-based action pruning and masking has been used to manage large action spaces in other RL domains, such as text-based games \cite{Ammanabrolu2020KGA2C}, illustrating that structural action constraints can be practically effective even when not motivated by pedagogical safety. Finally, student knowledge estimation in tutoring has long relied on probabilistic models such as
Bayesian Knowledge Tracing (BKT) \cite{CorbettAnderson1994BKT}, and subsequent
neural approaches such as Deep Knowledge Tracing \cite{Piech2015DKT} suggest
that richer representations of mastery may be possible, providing a
well-established family of mechanisms for representing evolving learner state
that can naturally interact with mastery-conditioned feasibility.

Taken together, existing work establishes strong foundations for constrained
optimization, safe exploration, and instructional RL. However,
mastery-conditioned feasibility where the admissible action set expands
endogenously as a function of learner knowledge appears to have received
limited formal treatment within the CMDP literature. This observation motivates
the development of algorithms that enforce pedagogical safety structurally,
rather than relying exclusively on reward engineering or post-hoc filtering.

For clarity, Table~\ref{tab:comparison} summarizes representative approaches
and highlights how MC-CPO differs in conditioning feasibility on learner
mastery and in providing structural pedagogical safety guarantees.


\begin{table}[!h]
\centering
\caption{Comparison of Safe RL and Educational RL Approaches}
\label{tab:comparison}
\footnotesize
\setlength{\tabcolsep}{3pt}
\begin{tabular}{lcccc}
\toprule
\textbf{Work} & \textbf{CMDP} & \textbf{Feasibility} & \textbf{Dynamic} & \textbf{Edu.} \\
\midrule
CPO \cite{Achiam2017CPO} & Yes & Static & No & No \\
RCPO \cite{Tessler2019RCPO} & Yes & Static & No & No \\
Lyapunov RL \cite{Chow2018Lyapunov} & Yes & Static & No & No \\
FPI \cite{Yang2023FPI} & Yes & Policy-based & Yes & No \\
CCPO \cite{Yao2023CCPO} & Yes & Threshold-cond. & No & No \\
KG-A2C \cite{Ammanabrolu2020KGA2C} & No & Graph mask & No & No \\
GITS \cite{Deng2023GITS} & No & Prerequisite & Implicit & Yes \\
Doroudi et al. \cite{Doroudi2019WheresReward} & No & N/A & N/A & Yes \\
\midrule
\textbf{MC-CPO (Ours)} & Yes & Mastery-cond. & Yes & Yes \\
\bottomrule
\end{tabular}
\end{table}


\section{Problem Setup}
\label{sec:problem_setup}

We formalize the instructional decision-making problem as a constrained Markov decision process (CMDP) in which feasibility is conditioned on learner mastery.

\subsection{Environment and Mastery State}

Let $\mathcal{V}$ denote a finite set of instructional concepts organized by a directed acyclic prerequisite graph $G = (\mathcal{V}, \mathcal{E})$, where $(u,v) \in \mathcal{E}$ indicates that concept $u$ is a prerequisite for concept $v$.

At time $t$, the environment state is defined as
\[
s_t = (x_t, K_t),
\]
where $x_t \in \mathcal{X}$ denotes observable learner interaction features, and $K_t \in [0,1]^{|\mathcal{V}|}$ is a mastery vector, with $K_t(v)$ representing the learner’s estimated mastery of concept $v$.

Mastery evolves according to a stochastic update rule
\[
K_{t+1} = \Phi(K_t, a_t, \xi_t),
\]
where $\xi_t$ represents stochastic learner response noise and $\Phi$ may correspond to Bayesian Knowledge Tracing (BKT) or another probabilistic mastery model \cite{CorbettAnderson1994BKT}.

\textbf{Assumption A1 (Bounded Mastery).}  
For all $t$ and $v \in \mathcal{V}$,
\[
0 \le K_t(v) \le 1.
\]
\textit{Pedagogical grounding:} This assumption is standard in probabilistic knowledge tracing models such as BKT \cite{CorbettAnderson1994BKT}, where mastery is modelled as a probability bounded in $[0,1]$. It holds by construction in any normalised competency model and does not require special environmental conditions.

The state transition kernel is denoted
\[
P(s_{t+1} \mid s_t, a_t).
\]

\subsection{Mastery-Conditioned Feasible Action Set}

Let $\mathcal{A}$ denote the full instructional action space, where each action corresponds to presenting or reinforcing a concept $v \in \mathcal{V}$.

We define the mastery-conditioned feasible set
\[
\mathcal{A}_f(s_t) = \left\{ a_v \in \mathcal{A} \,\middle|\, K_t(u) \ge \theta_{\min} \;\forall u \in \text{Pre}(v) \right\},
\]
where $\text{Pre}(v)$ denotes the prerequisite set of concept $v$, and $\theta_{\min} \in (0,1)$ is a fixed prerequisite mastery threshold.

Thus, an instructional action is admissible only if all prerequisite concepts exceed the mastery threshold. For example, a student who has not yet demonstrated sufficient mastery of algebra would not be presented calculus exercises, regardless of their potential engagement value.

\textbf{Assumption A2 (Non-emptiness).}  
For all reachable states $s_t$, $|\mathcal{A}_f(s_t)| \ge 1$.
\textit{Pedagogical grounding:} In any realistic curriculum, foundational concepts such as basic numeracy or literacy carry no prerequisites and are always admissible. Provided the prerequisite graph $G$ contains at least one source node (a concept with no incoming edges), A2 holds throughout training regardless of the learner's current mastery vector. Curriculum designers routinely satisfy this by construction.

Define the frontier set
\[
\mathcal{F}_t = \mathcal{A}_f(s_t) \setminus \mathcal{A}_f(s_{t-1}).
\]

Pedagogically, the frontier represents concepts that have just 
become accessible to a learner as prerequisite mastery is acquired the moment at which new instructional opportunities open.

\textbf{Proposition 1 (Monotonic Feasibility Expansion).}  
Under Assumption A1 and non-decreasing mastery updates in expectation,
\[
\mathbb{E}[K_{t+1}(v) \mid s_t, a_t] \ge K_t(v),
\]
the feasible set expands monotonically in expectation:
\[
\mathbb{E}[|\mathcal{A}_f(s_{t+1})|] \ge |\mathcal{A}_f(s_t)|.
\]

This expansion is state-driven via mastery progression rather than policy-driven via improved constraint compliance.

\subsection{Reward and Cost Structure}

We define an engagement reward function
\[
r_E : \mathcal{S} \times \mathcal{A} \to \mathbb{R}.
\]

Mastery does not appear directly in the reward; instead, it influences feasibility.

The objective is to maximize discounted engagement:
\[
J(\pi) = \mathbb{E}_\pi \left[ \sum_{t=0}^{\infty} \gamma^t r_E(s_t,a_t) \right].
\]

We define discounted cumulative cost functions corresponding to pedagogical safety constraints:
\[
J_{c_i}(\pi) = \mathbb{E}_\pi \left[ \sum_{t=0}^{\infty} \gamma^t c_i(s_t,a_t) \right], \quad i \in \{2,3,4\}.
\]

Each cost function captures violation of a pedagogical property:
\begin{itemize}
    \item $c_2(s_t,a_t)$: insufficient mastery progress,in instructional terms, this cost penalises episodes in which a learner is repeatedly exposed to content without measurable knowledge gain.
    \item $c_3(s_t,a_t)$: inadequate cognitive demand,this prevents the policy from defaulting to low-effort actions such as encouragement or hints that generate engagement without meaningful cognitive challenge.
    \item $c_4(s_t,a_t)$: engagement–learning decoupling.
\end{itemize}

The CMDP is therefore:
\[
\max_{\pi} J(\pi)
\]
subject to
\[
J_{c_i}(\pi) \le d_i, \quad i \in \{2,3,4\}.
\]
Budgets $d_i = \kappa_i \cdot J_{c_i}(\pi_{\mathrm{unc}})$ are set as
a fraction $\kappa_i \in (0,1)$ of the unconstrained baseline cost
$d_i^{\max} = J_{c_i}(\pi_{\mathrm{unc}})$, ensuring scale invariance
across environments. The unconstrained baseline cost $d_i^{\max}$ is
also used to normalize violation severity in the RHSI
(Section~\ref{sec:problem_setup}).

\subsection{Policy Class and Structural Safety}

We restrict the policy class to mastery-feasible actions via masked softmax parameterization:
\[
\pi_\theta(a \mid s_t)
=
\frac{\exp(f_\theta(s_t,a)) \mathbf{1}[a \in \mathcal{A}_f(s_t)]}
{\sum_{a' \in \mathcal{A}_f(s_t)} \exp(f_\theta(s_t,a')]}.
\]

\textbf{Assumption A3 (Mask Independence).}  
The feasibility indicator $\mathbf{1}[a \in \mathcal{A}_f(s)]$ depends only on state $s$ and is independent of policy parameters $\theta$.

\textbf{Lemma 1 (Structural Prerequisite Safety).}  
For all $\theta$ and all states $s_t$,
\[
\pi_\theta(a \mid s_t) = 0
\quad \text{if } a \notin \mathcal{A}_f(s_t).
\]

Thus, prerequisite violations are impossible by construction.

\subsection{Technical Assumptions}

We state additional regularity conditions under which the subsequent analysis holds.

The following regularity conditions support the subsequent analysis. \textbf{A4}: Rewards are bounded ($|r_E(s,a)| \le R_{\max}$). \textbf{A5}: Costs are bounded ($|c_i(s,a)| \le C_{\max}$). \textbf{A6}: State and action spaces are finite (adopted for theory; function approximation is used in experiments). \textbf{A7}: The executed policy satisfies $\tilde{\pi}(a|s) \ge \delta > 0$ for all $a \in \mathcal{A}_f(s)$; in neural implementations this is enforced approximately via entropy regularization and frontier mixing (Supplementary Appendix~D). \textbf{A8}: The prerequisite threshold $\theta_{\min}$ is fixed and independent of policy parameters. Under A1--A8, the mastery-conditioned CMDP is well-defined and admits stable primal--dual optimization.

\paragraph*{Reward Hacking Severity Index (RHSI).}
To quantify engagement learning decoupling, we define:
\[
\mathrm{RHSI}(\pi)
=
\frac{\hat{V}^\pi}{\hat{V}^*}
\times
\|\mathbf{v}(\pi)\|_w,
\]
where $\hat{V}^\pi = \mathbb{E}_\pi[\sum_t \gamma^t r_E(s_t,a_t)]$ is
the expected discounted engagement return under $\pi$,
$\hat{V}^* = \hat{V}^{\pi_{\mathrm{unc}}}$ is the unconstrained baseline
return, and
$\|\mathbf{v}(\pi)\|_w = \sqrt{\tfrac{1}{3}\sum_{i=2}^{4} v_i(\pi)^2}$
with $v_i(\pi) = J_{c_i}(\pi)/d_i^{\max}$ the normalized discounted
violation for constraint $C_i$ ($d_i^{\max}$ is the unconstrained
baseline cost). $\mathrm{RHSI}(\pi)$ is bounded in $[0,1]$: it equals
zero when constraints are fully satisfied or return collapses, and
increases as engagement and pedagogical violation rise simultaneously.
When $v_i(\pi) = J_{c_i}(\pi)/d_i^{\max}$ with $d_i^{\max} = J_{c_i}(\pi_{\mathrm{unc}})$,
the normalized form is bounded in $[0,1]$. Tables reporting \emph{RHSI (raw)}
use unnormalized discounted costs directly and are not subject to this bound;
they serve as a relative severity indicator across conditions within the
same experiment.

\textbf{Intuitive interpretation.} RHSI measures the joint severity of two failure modes that must co-occur for reward hacking to be present: high engagement return and high pedagogical constraint violation. A policy that achieves low return (e.g., a collapsed or cautious policy) will have low RHSI regardless of violations, since the engagement numerator suppresses the score. Conversely, a policy that satisfies all constraints will have low RHSI even at high return. RHSI is large only when the agent simultaneously maximises engagement \emph{and} violates safety constraints the precise pattern that constitutes reward hacking in instructional settings. A value of RHSI $\approx 0$ indicates safe behaviour; RHSI $\approx 1$ indicates maximal hacking under the normalisation. This self-contained definition applies uniformly across all experimental conditions reported in Section~\ref{sec:experiments}; no additional scaling or environment-specific calibration is required.

\section{MC-CPO Algorithm}
\label{sec:algorithm}

\begin{figure}[!h]
\centering
\includegraphics[width=\linewidth]{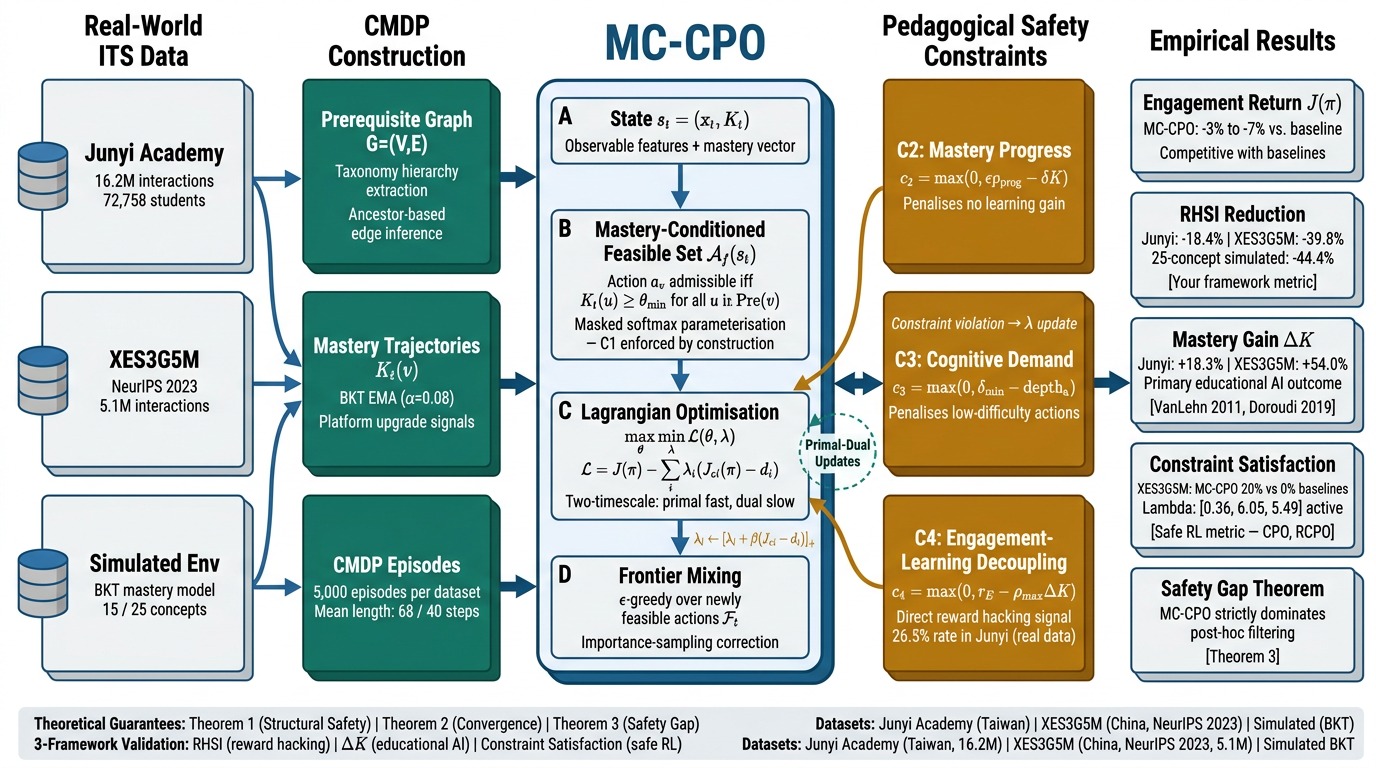}
\caption{MC-CPO pipeline. Real-world ITS data and a simulated BKT
environment are preprocessed into CMDP episodes with prerequisite
graphs and mastery trajectories. MC-CPO enforces mastery-conditioned
feasibility via masked softmax and optimises a primal--dual Lagrangian
with three pedagogical safety constraints. Results are validated under
three independent measurement frameworks: RHSI (reward hacking),
$\Delta K$ mastery gain (educational AI), and constraint satisfaction
rate (safe RL).}
\label{fig:architecture}
\end{figure}

We now present Mastery-Conditioned Constrained Policy Optimization (MC-CPO), a primal-dual policy-gradient method for solving the mastery-conditioned CMDP defined in Section~\ref{sec:problem_setup}. Figure~\ref{fig:architecture} provides an overview of the complete MC-CPO pipeline from data sources through algorithmic components to empirical evaluation.

\subsection{Lagrangian Formulation}

Define the Lagrangian:

\[
\mathcal{L}(\theta, \lambda)
=
J(\pi_\theta)
-
\sum_{i=2}^{4} \lambda_i \left( J_{c_i}(\pi_\theta) - d_i \right),
\]

where $\lambda_i \ge 0$ are dual variables associated with pedagogical safety constraints.

The primal objective is:

\[
\max_\theta \min_{\lambda \ge 0} \mathcal{L}(\theta, \lambda).
\]

Because prerequisite feasibility (C1) is enforced structurally via policy parameterization (Lemma 1), it does not appear in the Lagrangian.

\subsection{Policy Gradient Estimation}

Using standard policy-gradient theory, the gradient of the Lagrangian with respect to $\theta$ can be expressed as:

\[
\nabla_\theta \mathcal{L}
=
\mathbb{E}_\pi
\left[
\sum_{t=0}^\infty
\nabla_\theta \log \pi_\theta(a_t \mid s_t)
\left(
A_t^E
-
\sum_{i=2}^{4} \lambda_i A_t^{c_i}
\right)
\right],
\]

where:

- $A_t^E$ denotes the advantage estimator for engagement reward,
- $A_t^{c_i}$ denotes the advantage estimator for constraint cost $c_i$.

In practice, generalized advantage estimation (GAE) or Monte Carlo returns may be used.

Dual variables are updated via projected gradient ascent:

\[
\lambda_i \leftarrow
\left[
\lambda_i
+
\beta_k
\left( J_{c_i}(\pi_\theta) - d_i \right)
\right]_+.
\]

\subsection{Frontier-Conditioned Exploration}

When mastery increases, newly admissible actions in the frontier
$\mathcal{F}_t = \mathcal{A}_f(s_t) \setminus \mathcal{A}_f(s_{t-1})$
may initially receive low probability mass under $\pi_\theta$.
To mitigate potential probability collapse over newly feasible
actions, MC-CPO introduces event-triggered frontier mixing: when
$|\mathcal{F}_t| > 0$, the executed policy $\tilde{\pi}_{\theta,t}$
blends $\pi_\theta$ with a uniform distribution over $\mathcal{F}_t$
at rate $\epsilon_{\min}$, with importance-sampling correction to
preserve unbiased gradient estimation. This mechanism also provides
approximate practical enforcement of Assumption~A7. Its contribution
to asymptotic performance appears modest in the environments studied
(Supplementary Appendix~D), though it may provide greater benefit
in settings with longer prerequisite chains or sparser mastery
transitions.

\subsection{Two-Timescale Updates}
\label{sec:two_timescale}

We adopt a two-timescale stochastic approximation scheme:

\[
\theta_{k+1}
=
\theta_k
+
\alpha_k \nabla_\theta \mathcal{L}(\theta_k, \lambda_k),
\]

\[
\lambda_{k+1}
=
\left[
\lambda_k
+
\beta_k
\nabla_\lambda \mathcal{L}(\theta_k, \lambda_k)
\right]_+,
\]

with step sizes satisfying:

\[
\sum_k \alpha_k = \infty,
\quad
\sum_k \alpha_k^2 < \infty,
\quad
\sum_k \beta_k = \infty,
\quad
\sum_k \beta_k^2 < \infty,
\]
\[
\beta_k / \alpha_k \to 0.
\]

This ensures that primal updates occur on a faster timescale than dual updates.

\subsection{Algorithm Summary}

\begin{algorithm}[H]
\caption{MC-CPO: Mastery-Conditioned Constrained Policy Optimization}
\begin{algorithmic}[1]
\State Initialize $\theta$, $\lambda_2,\lambda_3,\lambda_4 \ge 0$
\For{each episode}
    \For{each step $t$}
        \State Observe $s_t = (x_t, K_t)$
        \State Compute feasible set $\mathcal{A}_f(s_t)$
        \State Compute frontier $\mathcal{F}_t$
        \State Set $\epsilon_t = \epsilon_{\min}\mathbf{1}[|\mathcal{F}_t|>0]$
        \State Sample $a_t \sim \tilde{\pi}_{\theta,t}$
        \State Observe $r_E$, $c_2$, $c_3$, $c_4$
        \State Update mastery $K_{t+1}$
    \EndFor
    \State Estimate returns and constraint costs
    \State Update $\theta$ via policy gradient
    \State Update $\lambda_i$ via projected ascent
\EndFor
\end{algorithmic}
\end{algorithm}


\section{Theoretical Guarantees}
\label{sec:theory}

Under Assumptions A1–A8, we establish structural prerequisite safety, convergence of the primal–dual updates, and a safety gap result comparing mastery-conditioned optimization to post-hoc filtering.

\subsection{Structural Prerequisite Safety}

We first formalize that prerequisite violations (C1) are impossible under the masked policy parameterization, independent of optimization dynamics.

\textbf{Theorem 1 (Structural Prerequisite Safety).}  
Under Assumptions A1–A3, for any policy parameters $\theta$ and any state $s_t$, the MC-CPO policy satisfies
\[
\pi_\theta(a \mid s_t) = 0
\quad
\text{whenever } a \notin \mathcal{A}_f(s_t).
\]
Consequently, the probability of prerequisite violation is identically zero for all $t$:
\[
\mathbb{P}_\pi(a_t \notin \mathcal{A}_f(s_t)) = 0.
\]

\textit{Proof.}  
By definition of the masked softmax parameterization,
\[
\pi_\theta(a \mid s_t)
=
\frac{\exp(f_\theta(s_t,a)) \mathbf{1}[a \in \mathcal{A}_f(s_t)]}
{\sum_{a' \in \mathcal{A}_f(s_t)} \exp(f_\theta(s_t,a')]}.
\]
If $a \notin \mathcal{A}_f(s_t)$, the indicator function evaluates to zero. Assumption A3 ensures that feasibility depends only on the state and is independent of $\theta$, so the mask cannot be circumvented by policy updates. Therefore,
\[
\pi_\theta(a \mid s_t) = 0
\]
for all infeasible actions and all parameter values. The result follows. \hfill $\square$


\subsection{Convergence of Primal–Dual Updates}

We now analyze convergence of the MC-CPO updates in the tabular setting under two-timescale stochastic approximation.

Let the Lagrangian be defined as
\[
\mathcal{L}(\theta,\lambda)
=
J(\pi_\theta)
-
\sum_{i=2}^4 \lambda_i \big( J_{c_i}(\pi_\theta) - d_i \big),
\]
with dual variables $\lambda_i \ge 0$.

The primal update is performed via stochastic gradient ascent:
\[
\theta_{k+1}
=
\theta_k
+
\alpha_k \widehat{\nabla_\theta \mathcal{L}}(\theta_k,\lambda_k),
\]
and the dual variables are updated via projected gradient ascent:
\[
\lambda_{i,k+1}
=
\Big[
\lambda_{i,k}
+
\beta_k \widehat{(J_{c_i}(\pi_{\theta_k}) - d_i)}
\Big]_+,
\]
for $i \in \{2,3,4\}$.

Step sizes satisfy the two-timescale conditions of Section~\ref{sec:two_timescale} \cite{Borkar2008SA}.

\medskip

\textbf{Theorem 2 (Convergence to Stationary Feasible Point).}  
Under Assumptions A1–A8, in the tabular setting, the MC-CPO iterates $(\theta_k,\lambda_k)$ converge almost surely to a stationary point $(\theta^\star,\lambda^\star)$ satisfying:

\begin{enumerate}
    \item Stationarity:
    \[
    \nabla_\theta \mathcal{L}(\theta^\star,\lambda^\star) = 0.
    \]
    
    \item Primal feasibility:
    \[
    J_{c_i}(\pi_{\theta^\star}) \le d_i,
    \quad
    i \in \{2,3,4\}.
    \]
    
    \item Complementary slackness:
    \[
    \lambda_i^\star \big(J_{c_i}(\pi_{\theta^\star}) - d_i\big) = 0.
    \]
\end{enumerate}

\textit{Proof Sketch.}  
Under Assumptions A4–A8, rewards and costs are bounded and the importance sampling ratios are uniformly bounded, implying bounded variance of gradient estimates. The masked policy parameterization satisfies Assumption A3, ensuring differentiability within the feasible set.

The primal–dual recursion constitutes a two-timescale stochastic approximation scheme. Standard results for two-timescale stochastic approximation specifically, the almost-sure convergence theorem established in Chapter~6 of \cite{Borkar2008SA} under Lipschitz and boundedness conditions satisfied by Assumptions A4--A8 imply that the fast-timescale parameter $\theta_k$ tracks the stationary points of the Lagrangian for quasi-static $\lambda$, while the slow-timescale dual variables ascend toward constraint satisfaction equilibria. Projected updates ensure non-negativity of $\lambda$ and enforce complementary slackness at equilibrium.

Because the feasible set $\mathcal{A}_f(s)$ depends only on the state and not on $\theta$, the stochastic approximation dynamics operate within a fixed parameterized policy class. Therefore, the iterates converge almost surely to a stationary saddle point of the Lagrangian. \hfill $\square$







\subsection{Safety Gap for Post-hoc Filtering}

We formalize post-hoc filtering as a two-stage procedure:
\begin{enumerate}
\item Learn an unconstrained policy
\[
\pi^{U} \in \arg\max_{\pi \in \Pi} J(\pi).
\]
\item At execution time, apply a state-dependent filter operator
\[
\mathcal{F} : \Pi \to \Pi_{\text{feas}}
\]
that enforces feasibility by modifying the action distribution only at states where infeasible actions are selected.
\end{enumerate}

We assume the filter preserves support over actions that the learned policy assigns nonzero probability, i.e.,
\[
\text{supp}(\mathcal{F}(\pi^U)(\cdot|s))
\subseteq
\text{supp}(\pi^U(\cdot|s)) \cup \mathcal{A}_f(s).
\]

\textbf{Theorem 3 (Safety Gap).}  
There exists a finite-horizon CMDP instance such that for any post-hoc filtering operator $\mathcal{F}$ satisfying the support-preservation condition above,

\[
J(\mathcal{F}(\pi^{U})) 
\;<\; 
\max_{\pi \in \Pi_{\text{feas}}} J(\pi).
\]

\textit{Proof sketch.}
Construct a horizon-2 CMDP with states $\{s_0, s_1, s_\bot\}$ and actions $\{a_\text{hack}, a_\text{prog}, a_\text{safe}\}$ at $s_0$. Set $r(s_0, a_\text{hack})=1$, $r(s_1, a_\text{learn})=R \in (0,1)$, all other rewards zero, and $c_2(s_0, a_\text{hack})=1$ with budget $d_2=0$. Unconstrained optimization yields $\pi^U(a_\text{hack}|s_0)=1$. The support-preservation condition then forces the filtered policy to assign zero mass to $a_\text{prog}$, yielding $J(\mathcal{F}(\pi^U))=0 < R = \max_{\pi \in \Pi_\text{feas}} J(\pi)$. The full CMDP construction and proof are given in Supplementary Appendix~A. \hfill $\square$

The result is most applicable when engagement incentives strongly dominate---precisely the reward hacking regime this paper targets.

Taken together, Theorems 1–3 establish that MC-CPO enforces prerequisite safety structurally, converges to stationary feasible points under standard stochastic approximation conditions, and can strictly outperform post-hoc filtering in terms of reward subject to safety constraints. These results provide formal justification for learning within the mastery-conditioned feasible set rather than correcting violations after unconstrained optimization. We now evaluate MC-CPO empirically in a simulated instructional environment.


\section{Experiments}
\label{sec:experiments}

This section evaluates whether MC-CPO enforces pedagogical safety while preserving engagement performance, and whether optimization within the mastery-conditioned feasible set yields improved reward–safety tradeoffs relative to post-hoc filtering.

All tabular experiments use 10 independent random seeds and 20,000 training episodes per method. Reported values are mean $\pm$ standard deviation across seeds. Final metrics are computed over the last 1,000 training episodes.

\subsection{Tabular Validation}

\subsubsection{Environment}

We construct a finite CMDP consistent with the theoretical formulation in Section~\ref{sec:problem_setup}.

The environment consists of:
\begin{itemize}
    \item $|\mathcal{V}| = 2$ instructional concepts arranged in a prerequisite chain of depth 1,
    \item two available actions at the initial state:
    \begin{itemize}
        \item \textbf{hack}: yields engagement reward $r_E = 1$ and terminates the episode, but incurs constraint cost $c = 1$,
        \item \textbf{prog}: yields reward $r_E = R$ with $R = 0.6$, satisfies all constraints ($c=0$), and produces mastery progression.
    \end{itemize}
\end{itemize}

Mastery progression is modeled as a Bernoulli update:
\[
\Delta K =
\begin{cases}
1 & \text{if prog is selected and learning succeeds}, \\
0 & \text{otherwise}.
\end{cases}
\]

Learning succeeds with probability 1 in the tabular setting, ensuring monotonic feasibility expansion consistent with Proposition 1.

The prerequisite threshold is $\theta_{\min} = 0.5$.
The discount factor is $\gamma = 0.99$.
Episodes terminate after a single action, making the environment analytically transparent.

This configuration matches the minimal construction used in the Safety Gap Theorem and ensures full reproducibility.

\subsubsection{Algorithms}

We compare:

\begin{itemize}
    \item \textbf{Unconstrained REINFORCE}
    \item \textbf{Reward-Shaped REINFORCE}
    \item \textbf{Post-hoc Filtering}
    \item \textbf{MC-CPO (Tabular)}
\end{itemize}

MC-CPO uses two-timescale primal–dual updates satisfying the step-size conditions of Theorem~2.

\textbf{Reward Shaping.}
Reward shaping augments the engagement reward with a linear penalty on learning-related costs:
\[
r^{\text{shape}}_t
=
r_E(t)
-
\alpha_2 c_2(t)
-
\alpha_4 c_4(t),
\]
where $\alpha_2, \alpha_4 > 0$ are fixed coefficients.

We use $\alpha_2 = 0.5$ and $\alpha_4 = 1.0$ across all experiments.
These coefficients were selected via a small pilot sweep and held fixed across all experiments to provide a consistent baseline. The 25-concept experiments additionally vary the shaping magnitude over $\lambda \in \{0.05, 0.1, 0.2, 0.5, 5.0\}$, 
finding no sensitivity to this choice at scale (Section~\ref{sec:25c}).

Because shaping optimizes a single scalar objective that trades reward against cost, it does not guarantee satisfaction of the discounted cost constraints. When engagement reward dominates the penalty terms, persistent reward hacking may remain optimal under the shaped objective.

\subsubsection{Training Dynamics}

\begin{figure}[t]
\centering
\includegraphics[width=\linewidth]{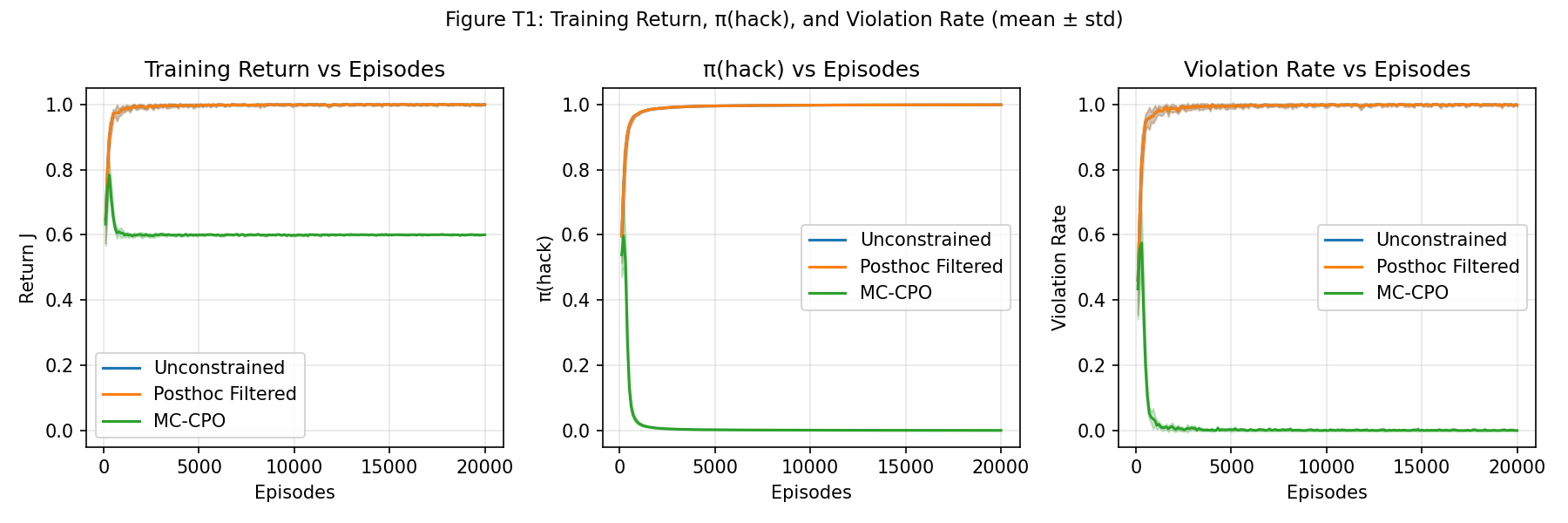}
\caption{Training return, $\pi(\text{hack})$, and violation rate (mean $\pm$ std across 10 seeds).}
\label{fig:tabular_training}
\end{figure}

Figure~\ref{fig:tabular_training} shows training return, $\pi(\text{hack})$, and violation rate.

Unconstrained training converges to $\pi(\text{hack}) \approx 1$, achieving near-maximal return ($0.999 \pm 0.001$) while violating constraints in nearly every episode.

Post-hoc filtering exhibits identical training behavior, as filtering is applied only during evaluation.

In contrast, MC-CPO drives $\pi(\text{hack})$ toward zero and maintains violation rate near zero after initial transients. Final engagement return converges to $0.600 \pm 0.001$, consistent with the theoretically expected return of the safe policy in this environment.

\subsubsection{Dual Convergence and Constraint Satisfaction}

\begin{figure}[t]
\centering
\includegraphics[width=\linewidth]{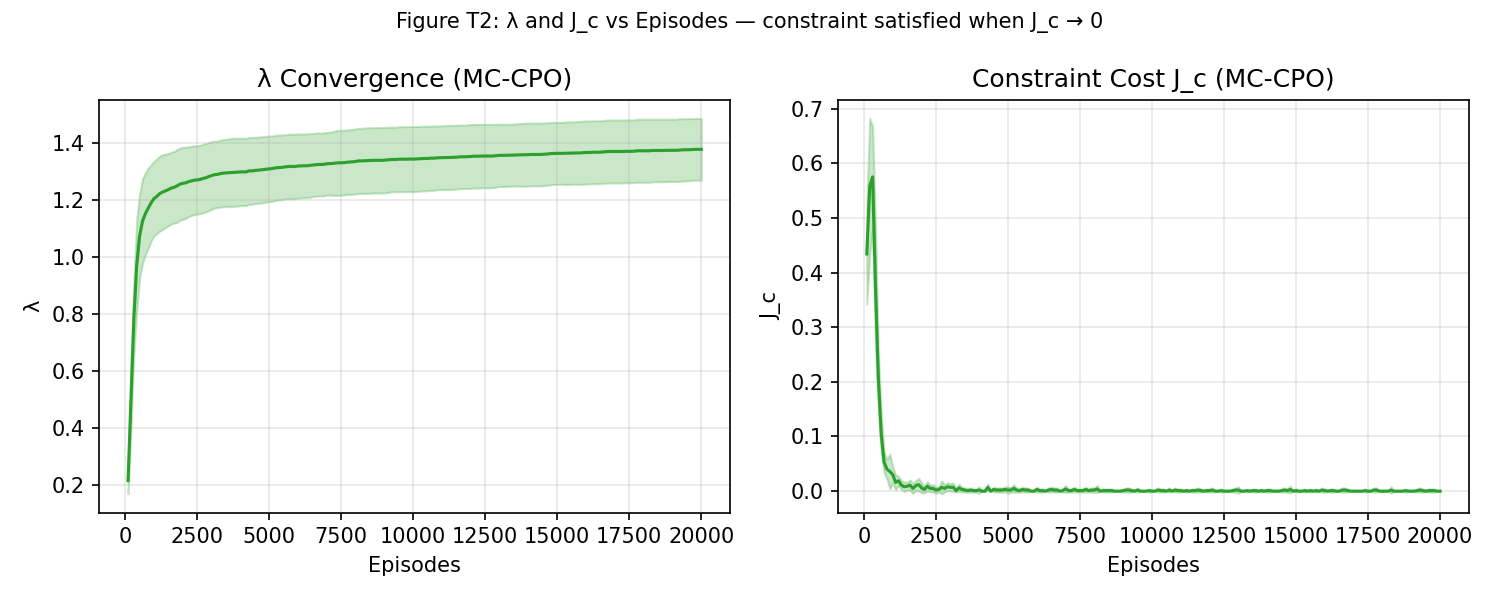}
\caption{Dual variable $\lambda$ and constraint cost $J_c$ during MC-CPO training.}
\label{fig:dual_convergence}
\end{figure}

Figure~\ref{fig:dual_convergence} reports the dual variable $\lambda$ and constraint cost $J_c$.

The dual variable increases during early violations and stabilizes once constraint satisfaction is achieved. The final constraint cost averaged over the last 1,000 episodes is:

\[
J_c^{\text{MC-CPO}} = 0.0007 \pm 0.0008.
\]

This empirically supports Theorem 1 and demonstrates stable primal–dual convergence.

\subsubsection{Safety Gap and RHSI}

Tables~\ref{tab:performance} and~\ref{tab:safety} summarize final performance and safety metrics.

\begin{table}[H]
\centering
\caption{Final performance (mean $\pm$ std; last 1,000 episodes).}
\label{tab:performance}
\begin{tabular}{lcc}
\toprule
Method & Return & $\pi(\text{hack})$ \\
\midrule
Unconstrained & $0.999 \pm 0.001$ & $0.999 \pm 0.000$ \\
Post-hoc & $0.0005 \pm 0.0006$ & $0.999 \pm 0.000$ \\
MC-CPO & $0.600 \pm 0.001$ & $0.0004 \pm 0.0001$ \\
\bottomrule
\end{tabular}
\end{table}

\begin{table}[H]
\centering
\caption{Safety and reward hacking severity (mean $\pm$ std; 
last 1,000 episodes).}
\label{tab:safety}
\begin{tabular}{lccc}
\toprule
Method & $J_c$ & Viol. Rate & RHSI \\
\midrule
Unconstrained    & $0.998 \pm 0.001$ & $0.998 \pm 0.001$ & $0.998 \pm 0.001$ \\
Post-hoc (train) & $0.000 \pm 0.000$ & $0.000 \pm 0.000$ & $0.998 \pm 0.001$ \\
MC-CPO           & $0.0007 \pm 0.0008$ & $0.0007 \pm 0.0008$ & $0.0004 \pm 0.0001$ \\
\bottomrule
\end{tabular}
\end{table}

RHSI is computed as defined in Section~\ref{sec:problem_setup}.
Unconstrained and post-hoc methods achieve RHSI $\approx 0.998$,
reflecting high engagement return combined with near-total constraint
violation. MC-CPO reduces RHSI to $0.0004 \pm 0.0001$, indicating
that constraint satisfaction and the engagement--mastery tradeoff
are achieved simultaneously.

MC-CPO achieves near-zero violation while preserving substantially
higher return than post-hoc filtering.

A Welch t-test comparing MC-CPO and post-hoc filtered return yields:
\[
t = 2438.68, \quad p < 0.01.
\]
The effect size is large (Cohen's $d = 1149.60$). This unusually
high value reflects the structural separation between the policies:
post-hoc filtering collapses return toward zero while MC-CPO
converges deterministically to the safe optimum rather than
exhibiting distributional overlap.

These results empirically validate the Safety Gap Theorem:
optimizing within the mastery-conditioned feasible set strictly
dominates post-hoc filtering under equal safety constraints.

\begin{figure}[H]
\centering
\includegraphics[width=\linewidth]{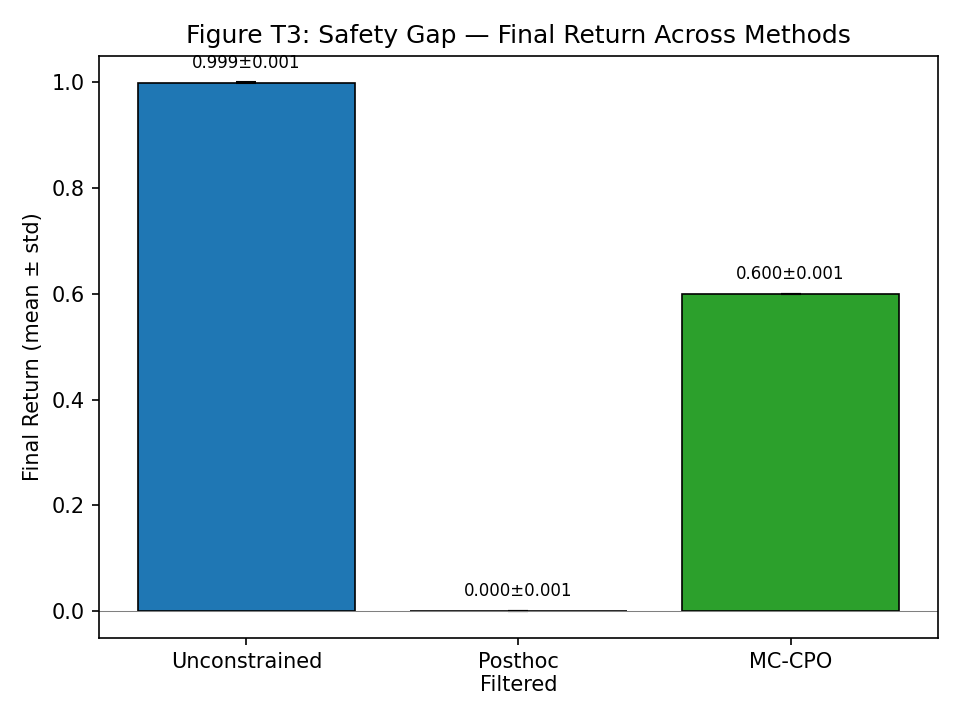}
\caption{Final return across methods (mean $\pm$ std across 10 seeds).}
\label{fig:safety_gap}
\end{figure}

\subsubsection{Robustness to Reward Gap}

We evaluate robustness across reward magnitudes $R \in \{0.5, 0.6, 0.7\}$.

Across all settings:
\begin{itemize}
    \item $\pi(\text{hack}) < 5 \times 10^{-4}$,
    \item $J_c \approx 7 \times 10^{-4}$,
    \item RHSI remains bounded near $R$,
    \item Final return scales proportionally with $R$.
\end{itemize}

These results indicate that reward hacking elimination is structural rather than tuned to a specific reward scale.

\subsubsection{Extended Tabular Validation (Multi-Step Stochastic CMDP)}
\label{sec:tabular_extended}

To evaluate robustness beyond the single-step minimal construction, we consider a multi-step tabular CMDP with stochastic mastery transitions and dynamic feasibility expansion.

\paragraph{Environment.}
The environment contains $|\mathcal{V}| = 5$ instructional concepts arranged in a prerequisite chain:
\[
1 \rightarrow 2 \rightarrow 3 \rightarrow 4 \rightarrow 5.
\]
Episodes have horizon $T = 5$ and discount factor $\gamma = 0.99$.
The state is $s_t = (K_t, t)$ where $K_t \in \{0,1\}^5$ denotes binary mastery.

The feasible action set expands with mastery:
\[
\mathcal{A}_f(s_t) = \{1\} \cup \{ i \in \{2,\dots,5\} : K_{t,i-1}=1 \}.
\]

A special action \textbf{hack} remains available at all timesteps and yields engagement reward $r_E = 1.0$ but produces no mastery progression.
Selecting a feasible concept yields engagement reward $r_E = 0.6$ and increases mastery stochastically with probability $p_{\mathrm{learn}} = 0.8$ when the concept is not yet mastered.

Mastery change per step is:
\[
\Delta K_t = \sum_{i=1}^5 (K_{t+1,i} - K_{t,i}).
\]

\paragraph{Costs.}
We define three discounted costs:
\begin{itemize}
    \item $c_2(t) = \mathbf{1}[a_t \in \mathcal{A}_f(s_t),\, a_t \neq \text{hack},\, \Delta K_t = 0]$,
    \item $c_3(t) = \mathbf{1}[a_t \notin \mathcal{A}_f(s_t)]$,
    \item $c_4(t) = \mathbf{1}[r_E(t) > 0,\, \Delta K_t = 0]$.
\end{itemize}

\paragraph{Protocol.}
All methods run for 200{,}000 episodes (10 seeds); reported values are mean $\pm$ std over seeds, computed over the final 1{,}000 episodes. All methods share identical budgets, discount factor, learning rate, and seeds the sole variation is the optimisation objective. Standard safe RL algorithms (CPO~\cite{Achiam2017CPO}, RCPO~\cite{Tessler2019RCPO}) require learner-state-independent cost limits, which defeats the purpose of mastery-conditioned feasibility; the chosen baselines differ from MC-CPO only in how safety is enforced.

\paragraph{Main results.}

Performance metrics are reported in Table~\ref{tab:chain_performance}, and safety metrics in Table~\ref{tab:chain_safety}.

Unconstrained training converges to a high hacking probability and extremely large RHSI, indicating engagement gains obtained without corresponding mastery improvement. The extreme variance in unconstrained raw RHSI ($\sigma > \mu$) reflects near-zero mastery gain in some seeds, where the ratio form approaches instability; this motivates the product-form RHSI adopted in the neural experiments.

Post-hoc filtering eliminates hacking at evaluation time but incurs a persistent reward gap.  
MC-CPO reduces $\pi(\text{hack})$ while maintaining substantially higher return than post-hoc filtering, and yields lower RHSI, consistent with optimizing inside the mastery-conditioned feasible set.

\begin{table}[H]
\centering
\caption{Extended chain CMDP performance (mean $\pm$ std; last 1{,}000 episodes; 10 seeds). RHSI reported in raw form; the normalized form (Section~\ref{sec:problem_setup}) applies where violation rates are available.}
\label{tab:chain_performance}
\begin{tabular}{lccc}
\toprule
Method & Return & $\pi(\text{hack})$ & RHSI (raw) \\
\midrule
Unconstrained
& $4.768 \pm 0.215$
& $0.932 \pm 0.110$
& $(2.45 \pm 2.58)\times10^{6}$ \\

Post-hoc
& $2.941 \pm 0.000$
& $0.000 \pm 0.000$
& $2.406 \pm 0.508$ \\

MC-CPO
& $3.538 \pm 0.553$
& $0.304 \pm 0.281$
& $2.179 \pm 1.909$ \\
\bottomrule
\end{tabular}
\end{table}

\begin{table}[t]
\centering
\caption{Extended chain CMDP safety metrics (mean $\pm$ std; last 1{,}000 episodes; 10 seeds).}
\label{tab:chain_safety}
\begin{tabular}{lcc}
\toprule
Method & $J_{c_2}$ & $J_{c_4}$ \\
\midrule
Unconstrained
& $0.069 \pm 0.111$
& $4.637 \pm 0.427$ \\

Post-hoc
& $3.626 \pm 0.326$
& $3.626 \pm 0.326$ \\

MC-CPO
& $1.098 \pm 0.498$
& $2.592 \pm 0.993$ \\
\bottomrule
\end{tabular}
\end{table}

\begin{figure}[t]
\centering
\includegraphics[width=\linewidth]{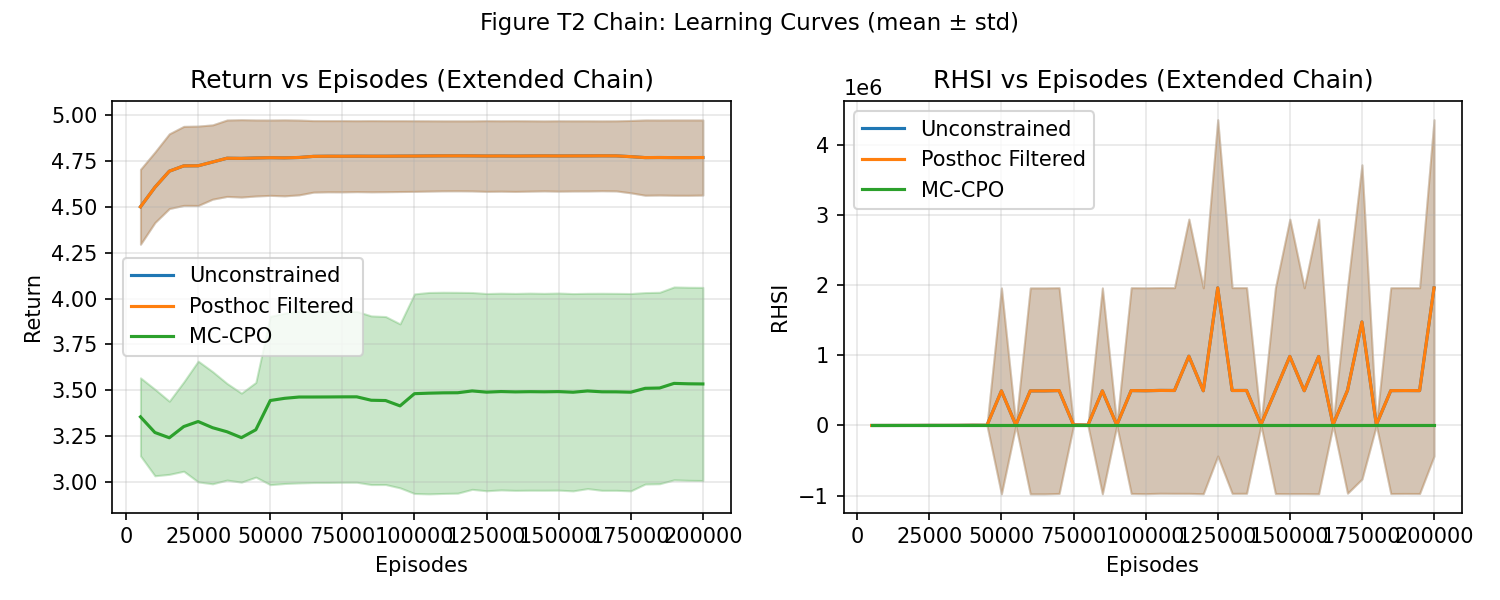}
\caption{Extended chain CMDP learning curves (mean $\pm$ std across 10 seeds). Left: return vs episodes. Right: RHSI vs episodes.}
\label{fig:t2_chain_learning}
\end{figure}

\paragraph{Interpretation.}
The extended chain stress-tests the safety gap mechanism under multi-step interaction and stochastic mastery.

Unconstrained policies tend to select \textbf{hack} predominantly, which drives raw RHSI to extreme values reflecting near-zero mastery gain relative to engagement return.
Post-hoc filtering prevents hacking at evaluation time but does not optimize for feasible progression, resulting in a persistent reward gap.
MC-CPO appears to improve the reward--learning balance by optimizing under mastery-conditioned feasibility and discounted costs, yielding a higher-return safe policy than post-hoc filtering, though variability across seeds suggests sensitivity to initialization in this stochastic setting.


\subsection{Neural Policy Implementation}
\label{sec:neural}

We next evaluate MC-CPO under function approximation using a PPO backbone to assess whether the theoretical guarantees observed in the tabular regime extend to neural policies.

\subsubsection{Simulation Environment}

We simulate an instructional environment with $|\mathcal{V}| = 15$ concepts. Mastery evolves according to a Bayesian Knowledge Tracing (BKT) update with learning rate $\eta = 0.08$. A concept becomes eligible once mastery exceeds a prerequisite threshold $\theta_{\min} = 0.7$.

Engagement reward $r_E(s,a)$ combines base engagement, novelty bonus, and difficulty penalty. The reward is bounded and satisfies Assumption A4.

Episodes have horizon $T=100$ steps and are evaluated over 200 episodes per seed across 10 seeds.

\subsubsection{Protocol and Baselines}

We compare Unconstrained PPO, Reward-Shaped PPO ($r^\text{shape}_t = r_E(t) - 0.5c_2(t) - c_4(t)$), Post-hoc Filtered PPO, MC-CPO, and MC-CPO without frontier mixing---all using identical two-layer MLP policies (hidden sizes $(64,64)$) and optimizer settings. Post-hoc filtering applies evaluation masking without modifying training, yielding identical asymptotic metrics to the unconstrained baseline. Constraint budgets are $d_i = \kappa_i \bar{J}_{c_i}^U$ relative to unconstrained costs; a constraint is satisfied if $\bar{J}_{c_i} \le (1+\tau)d_i$, $\tau=0.1$.

\subsubsection{Main Results}

Table~\ref{tab:neural_results} summarizes final performance under neural function approximation. Results are reported as mean $\pm$ standard deviation across 10 seeds, with each seed evaluated over 200 episodes.

\begin{table}[H]
\centering
\caption{Neural final evaluation results (mean $\pm$ std over 10 seeds; RHSI reported in raw form). Constraint satisfaction: $\bar{J}_{c_i} \le (1+\tau)d_i$, $\tau = 0.1$.}
\label{tab:neural_results}
\resizebox{\columnwidth}{!}{
\begin{tabular}{lccccc}
\toprule
Method & Return & RHSI (raw) & $J_{c_2}$ & $J_{c_3}$ & $J_{c_4}$ \\
\midrule
Unconstrained 
& 34.64 $\pm$ 0.37 
& 69.92 $\pm$ 0.85 
& 0.917 
& 19.02 
& 27.14 \\

Reward-Shaped 
& 34.64 $\pm$ 0.37 
& 69.92 $\pm$ 0.85 
& 0.917 
& 19.02 
& 27.14 \\

Post-hoc 
& 34.64 $\pm$ 0.37 
& 69.92 $\pm$ 0.85 
& 0.917 
& 19.02 
& 27.14 \\

\textbf{MC-CPO} 
& \textbf{32.73 $\pm$ 0.48} 
& \textbf{44.47 $\pm$ 0.96} 
& \textbf{0.871} 
& \textbf{9.69} 
& \textbf{23.14} \\

MC-CPO (no frontier) 
& 32.68 $\pm$ 0.39 
& 44.51 $\pm$ 0.81 
& 0.872 
& 9.69 
& 23.12 \\
\bottomrule
\end{tabular}
}
\end{table}

Unconstrained, reward-shaped, and post-hoc methods achieve identical engagement return ($34.64 \pm 0.37$) and high raw RHSI ($69.92 \pm 0.85$), suggesting systematic engagement without proportional mastery gains. In the neural setting, reward shaping introduces a small additive penalty relative to engagement reward; diagnostic checks confirm shaping is active but appears to produce negligible long-run deviation in this environment. Post-hoc filtering shares the same evaluation masking as the unconstrained baseline, yielding identical final metrics.

MC-CPO reduces engagement modestly ($32.73 \pm 0.48$) while substantially lowering raw RHSI ($44.47 \pm 0.96$), suggesting improved alignment between engagement and mastery improvement. Discounted costs are reduced across all three constraints:
\[
J_{c_2} = 0.871, \quad
J_{c_3} = 9.69, \quad
J_{c_4} = 23.14,
\]
all within the $(1+\tau)d_i$ thresholds under the stated tolerance ($\tau = 0.1$), indicating that MC-CPO approximately satisfies all pedagogical safety constraints under neural approximation.

\begin{figure}[H]
\centering
\includegraphics[width=\columnwidth]{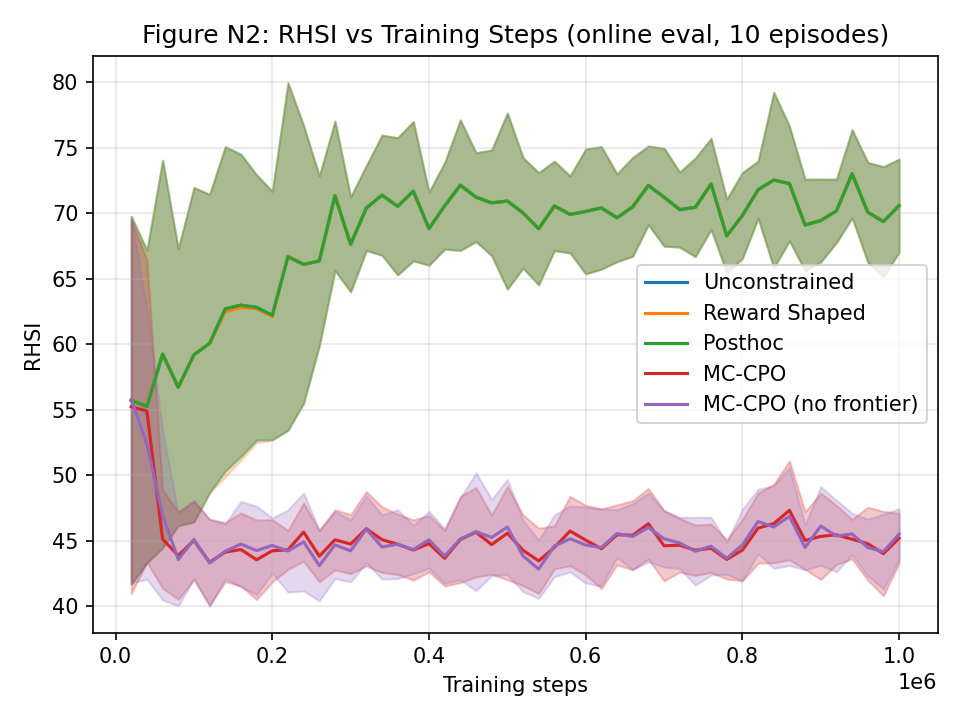}
\caption{RHSI over training steps (15-concept simulated environment, mean $\pm$ std across 10 seeds). MC-CPO is the only method to reduce RHSI below the unconstrained baseline.}
\label{fig:n2_rhsi}
\end{figure}

Figure~\ref{fig:n2_rhsi} reports RHSI trajectories over training. Return, cost, and frontier ablation curves are provided in Supplementary Appendix~B.

Overall, MC-CPO achieves an improved reward--safety tradeoff relative to baseline PPO variants, reducing reward hacking severity while maintaining competitive engagement performance.

\subsection{Scaling to Twenty-Five Concepts}
\label{sec:25c}

We extend the neural environment to $|\mathcal{V}|=25$ concepts,
representative of realistic tutoring curricula with deep prerequisite
graphs, to investigate whether MC-CPO's structural guarantees may
hold at this scale. Budgets are re-anchored from unconstrained
baseline costs at this scale. Table~\ref{tab:25c} reports results
over 10 seeds and 1M steps.

Among the methods evaluated, MC-CPO is the only one that separates
from the unconstrained baseline, reducing raw RHSI from $70.52$ to
$39.16$ while achieving a constraint satisfaction rate of 80\%
(8 of 10 seeds). The two unsatisfied seeds violate the cognitive
demand constraint ($C_3$) by small margins, while satisfying $C_2$
and $C_4$ in all cases. Unconstrained, posthoc, and reward-shaped
policies satisfy no constraint budgets across any seed.

\paragraph*{Reward shaping at scale.}
Reward shaping with $\lambda \in \{0.05, 0.1, 0.2, 0.5, 5.0\}$
produces policies indistinguishable from the unconstrained baseline
(return $34.79$, raw RHSI $70.52$, constant across all $\lambda$).
At 25-concept scale, per-step mastery increments $\Delta K_t \approx 0$
because the prerequisite graph depth creates a sparse learning signal;
the shaping term $\lambda \cdot \Delta K_t$ consequently contributes
near-zero gradient regardless of $\lambda$.
These results suggest that additive penalty methods may be structurally
ineffective in the sparse-learning regime that characterises realistic
tutoring deployments at scale a setting where MC-CPO's structural
constraint approach appears to remain effective.

\begin{table}[H]
\centering
\caption{25-concept neural results (10 seeds, 1M steps). Raw RHSI 
reported. Constraint satisfaction rate reflects seeds satisfying 
all three budgets within tolerance $\tau=0.1$.}
\label{tab:25c}
\footnotesize
\setlength{\tabcolsep}{3pt}
\begin{tabular}{lccc}
\toprule
\textbf{Method} & \textbf{Return} & \textbf{RHSI (raw)} 
& \textbf{Sat. Rate} \\
\midrule
Unconstrained        & $34.79 \pm 0.30$ & $70.52 \pm 0.68$ & 0\% \\
Reward Shaped (all $\lambda$) & $34.79 \pm 0.30$ & $70.52 \pm 0.68$ & 0\% \\
Posthoc              & $34.79 \pm 0.30$ & $70.52 \pm 0.68$ & 0\% \\
\midrule
\textbf{MC-CPO}      & $\mathbf{31.97 \pm 0.31}$ & $\mathbf{39.16 \pm 2.02}$ & \textbf{80\%} \\
MC-CPO (no frontier) & $32.11 \pm 0.52$ & $40.05 \pm 2.93$ & 70\% \\
\bottomrule
\end{tabular}
\end{table}

\subsection{Real-World Validation on Public ITS Datasets}
\label{sec:real_world}

To evaluate whether the reward hacking pattern modelled by MC-CPO is observable in deployed ITS systems, and whether MC-CPO reduces it on real student data, we conduct experiments on two publicly available datasets.

\subsubsection{Datasets and Preprocessing}

\paragraph{Junyi Academy.}
The Junyi Academy dataset~\cite{JunyiAcademy2015} contains 16{,}217{,}311 student--exercise interaction logs from 72{,}758 students across 1{,}330 mathematics exercises from a deployed Taiwanese e-learning platform. Each record includes correctness (\texttt{is\_correct}), platform-assessed mastery changes (\texttt{is\_upgrade}, \texttt{is\_downgrade}), hint usage, and timestamps. The platform's upgrade and downgrade signals provide a ground-truth mastery oracle requiring no separately fitted knowledge-tracing model.

A prerequisite graph is extracted from the four-level curriculum taxonomy (\texttt{level1\_id}--\texttt{level4\_id}) and difficulty annotations. Within each topic cluster, edges follow easy~$\rightarrow$~medium~$\rightarrow$~hard ordering; cross-stage edges link elementary to junior concepts within shared branches. This yields $|\mathcal{V}|=14$ concepts and 17 prerequisite edges. Mastery trajectories $K_t(v)$ are estimated via exponential moving average of the platform's upgrade/downgrade signals ($\alpha=0.08$). After filtering students with fewer than 30 interactions, 5{,}000 episodes are retained (mean length $68.2$ steps).

\paragraph{XES3G5M.}
XES3G5M is a knowledge tracing benchmark introduced at NeurIPS 2023~\cite{Liu2023XES3G5M}, containing 5{,}139{,}044 interactions from 14{,}453 students across 865 knowledge components (KCs) and 7{,}652 exercises on a Chinese mathematics tutoring platform. The dataset provides KC-level learning routes (\texttt{kc\_routes}) encoding full prerequisite hierarchies as delimited paths (e.g., \texttt{A\texttt{----}B\texttt{----}C\texttt{----}D}).

Ancestor-based prerequisite edges are extracted: KC $A$ is a structural prerequisite of KC $B$ if $A$ appears before $B$ in any shared route path. Restricting to the top 100 KCs by interaction frequency and filtering for at least one ancestor edge yields $|\mathcal{V}|=14$ concepts and 12 prerequisite edges. Mastery is estimated via BKT EMA on correctness ($\alpha=0.08$). After filtering students with fewer than 20 interactions, 5{,}000 episodes are retained (mean length $39.9$ steps).

\subsubsection{Reward Hacking Prevalence in Deployed Data}

Before training MC-CPO, we quantify the $C_4$ (engagement--learning decoupling) event rate directly in the raw interaction logs. A $C_4$ event is recorded when a student answers correctly (high engagement signal) but the platform does not register a mastery upgrade (no real learning gain).

On Junyi Academy, 90{,}485 of 340{,}988 analysed interactions (\textbf{26.5\%}) exhibit this pattern. On XES3G5M, 6{,}269 of 199{,}605 interactions (\textbf{3.1\%}) exhibit this pattern. The difference reflects student population characteristics: Junyi students exhibit a mean correctness rate of 65\%, generating a wider engagement--mastery gap, while XES3G5M students exhibit 83\% correctness. Both rates confirm that the reward hacking pattern targeted by MC-CPO is directly observable in deployed ITS data at scale, and not a simulation artifact.

\subsubsection{Experimental Protocol}

All five methods (Unconstrained PPO, Reward-Shaped PPO, Post-hoc Filtered PPO, MC-CPO, MC-CPO without frontier mixing) are trained using the \texttt{JunyiTutoringEnv} environment wrapper, which loads the preprocessed prerequisite graph and computes mastery-conditioned feasibility sets identically to the simulated setting. All hyperparameters match the 15-concept neural experiment: hidden sizes $(64,64)$, $\eta=0.08$, $\theta_{\min}=0.7$, $\gamma=0.99$, 10 seeds, $1\times10^6$ training steps, 200 final evaluation episodes per seed.

Constraint budgets are anchored at $\kappa=0.90$ of unconstrained baseline costs: $d_i = 0.90 \cdot J_{c_i}(\pi_{\mathrm{unc}})$. Setting $\kappa < 1$ ensures $J_{c_i}^{\mathrm{unc}} > d_i$, guaranteeing that dual variables activate and methods genuinely differentiate---a stronger test than the $\kappa=1.02$ used in the simulated setting, where baseline costs are more stable.

\subsubsection{Results}

Table~\ref{tab:real_world} reports mean final evaluation results across 10 seeds.

\begin{table}[H]
\centering
\caption{Real-world validation results (mean across 10 seeds; 200 final evaluation episodes per seed). RHSI reported in raw form. Sat.: fraction of seeds satisfying all three budget constraints within tolerance $\tau=0.1$. NF: no frontier mixing.}
\label{tab:real_world}
\resizebox{\columnwidth}{!}{%
\begin{tabular}{llccccc}
\toprule
\textbf{Dataset} & \textbf{Method} & \textbf{Return} & \textbf{RHSI} & $J_{c_2}$ & $J_{c_3}$ & $J_{c_4}$ \\
\midrule
\multirow{5}{*}{\shortstack[l]{\textbf{Junyi Academy}\\(16.2M interactions)\\14 concepts, 17 edges\\Mastery: platform signal}}
& Unconstrained   & 30.21 & 57.29 & 0.674 & 12.378 & 23.920 \\
& Reward-Shaped   & 30.21 & 57.29 & 0.674 & 12.378 & 23.920 \\
& Post-hoc        & 30.21 & 57.29 & 0.674 & 12.378 & 23.920 \\
& \textbf{MC-CPO} & \textbf{29.27} & \textbf{46.74} & \textbf{0.707} & \textbf{12.378} & \textbf{22.283} \\
& MC-CPO (NF)     & 29.27 & 46.74 & 0.707 & 12.378 & 22.284 \\
\midrule
\multirow{5}{*}{\shortstack[l]{\textbf{XES3G5M}\\(NeurIPS 2023, 5.1M)\\14 concepts, 12 edges\\Mastery: BKT on correctness}}
& Unconstrained   & 34.64 & 69.92 & 0.917 & 19.019 & 27.138 \\
& Reward-Shaped   & 34.64 & 69.92 & 0.917 & 19.019 & 27.138 \\
& Post-hoc        & 34.64 & 69.92 & 0.917 & 19.019 & 27.138 \\
& \textbf{MC-CPO} & \textbf{32.26} & \textbf{42.11} & \textbf{0.877} & \textbf{9.693} & \textbf{22.564} \\
& MC-CPO (NF)     & 32.38 & 42.36 & 0.876 &  9.693 & 22.697 \\
\bottomrule
\end{tabular}
}
\end{table}

On both datasets, unconstrained, reward-shaped, and post-hoc methods converge to identical engagement returns and RHSI values, consistent with findings in the simulated neural setting (Section~\ref{sec:neural}). Reward shaping produces negligible RHSI reduction: mastery increments per step are small, making the shaping gradient $\lambda \cdot \Delta K_t$ near-zero regardless of $\lambda$. Post-hoc filtering shares the same evaluation masking as the unconstrained baseline, yielding identical final metrics.

MC-CPO is the only method that separates from the unconstrained baseline on both datasets. On Junyi Academy, MC-CPO activates non-zero Lagrange multipliers ($\hat{\lambda} = [0.356, 6.053, 5.485]$) and reduces raw RHSI by \textbf{18.4\%} ($57.29 \rightarrow 46.74$) while maintaining competitive return ($-3.1\%$ relative). The $C_4$ discounted cost falls from 23.92 to 22.28 ($-6.9\%$), directly reducing engagement--learning decoupling. On XES3G5M, dual variables activate ($\hat{\lambda} = [0.377, 7.714, 8.753]$) and RHSI falls by \textbf{39.8\%} ($69.92 \rightarrow 42.11$) with a $-7.0\%$ return cost; $J_{c_4}$ falls from 27.14 to 22.56 ($-16.8\%$). Constraint satisfaction (at $\tau=0.1$) is achieved in 20\% of seeds for MC-CPO vs.\ 0\% for all baselines on XES3G5M, and in 0\% vs.\ 0\% on Junyi Academy (where $\kappa=0.90$ yields especially tight budgets).

The stronger RHSI reduction on XES3G5M ($39.8\%$) relative to Junyi Academy ($18.4\%$) is consistent with the higher unconstrained RHSI baseline on that dataset ($69.92$ vs.\ $57.29$), reflecting a wider engagement--mastery gap available for the Lagrangian to correct.

\paragraph{Mastery gain validation ($\Delta K$).}
Beyond RHSI, we report mean per-episode mastery gain $\Delta K$ as an independent, externally grounded outcome measure aligned with the educational AI literature~\cite{VanLehn2011ITS,Doroudi2019WheresReward,Rafferty2011POMDP}. This metric requires no reference to the RHSI definition and directly quantifies the pedagogical benefit of each policy in terms of actual knowledge acquisition.

The results provide convergent evidence for MC-CPO's effectiveness. On Junyi Academy, MC-CPO increases mean mastery gain from $\Delta K = 0.729$ (unconstrained) to $\Delta K = 0.862$, a \textbf{+18.3\% improvement} in knowledge acquisition per episode. On XES3G5M, the improvement is more substantial: $\Delta K$ rises from $0.796$ to $1.226$, a \textbf{+54.0\% improvement}. In both cases, unconstrained, reward-shaped, and post-hoc methods produce identical $\Delta K$ values, confirming they do not promote additional learning relative to each other.

These $\Delta K$ improvements are not explained by return differences alone---MC-CPO achieves lower engagement return than baselines while simultaneously producing substantially higher mastery gain. This dissociation is precisely the pedagogical safety property that MC-CPO is designed to enforce: the structural constraint separates engagement optimization from learning outcomes, redirecting policy behaviour toward concept sequences that produce genuine knowledge gains. Taken together, the convergent reduction in RHSI and increase in $\Delta K$ across two independently collected datasets provide multi-metric, multi-tradition evidence that MC-CPO's structural constraint approach generalises beyond the simulated setting.

\begin{figure}[H]
\centering
\includegraphics[width=\columnwidth]{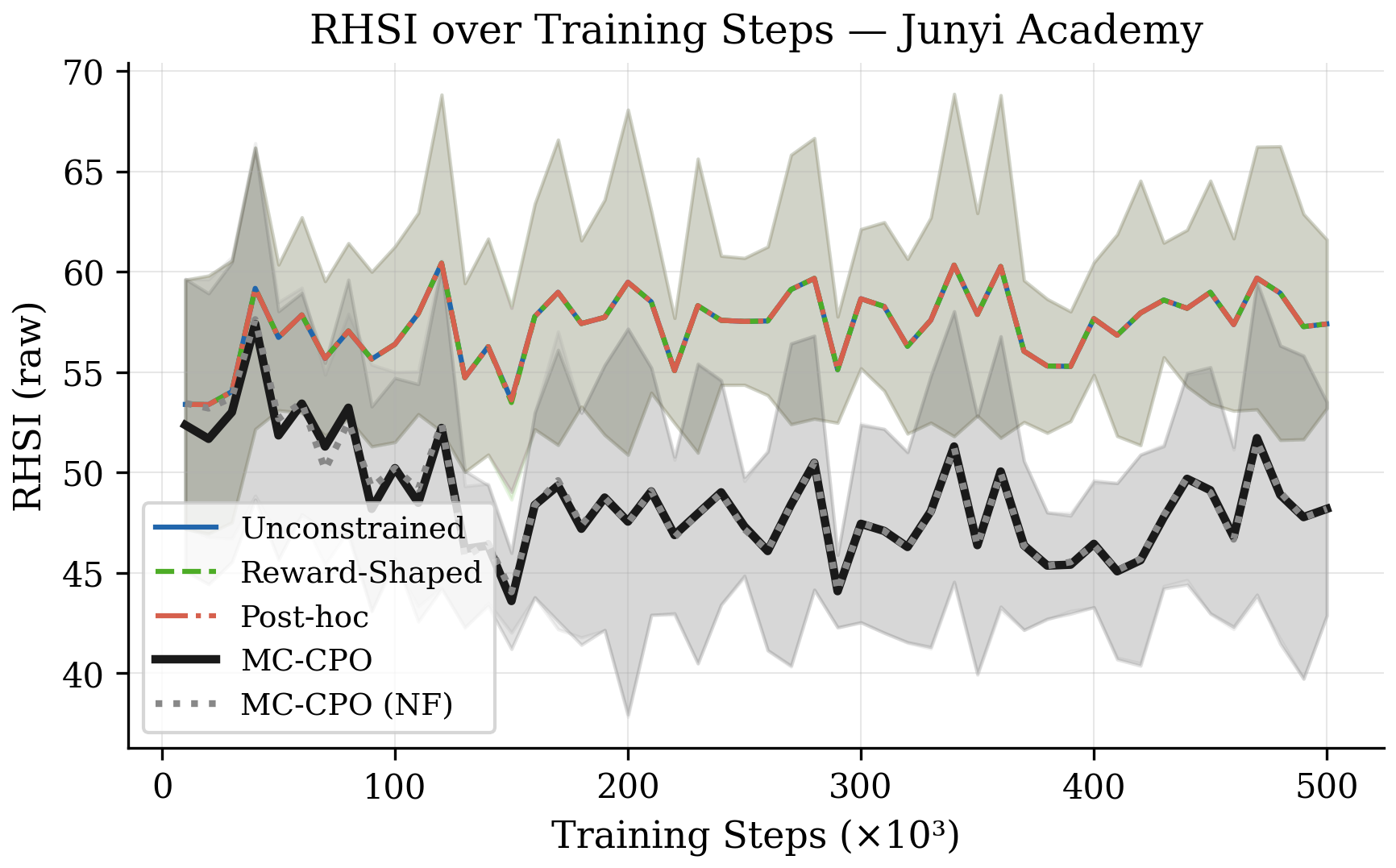}
\caption{RHSI over training steps on Junyi Academy (mean $\pm$ std, 10 seeds). MC-CPO is the only method to reduce RHSI below the unconstrained baseline.}
\label{fig:junyi_rhsi}
\end{figure}

\begin{figure}[H]
\centering
\includegraphics[width=\columnwidth]{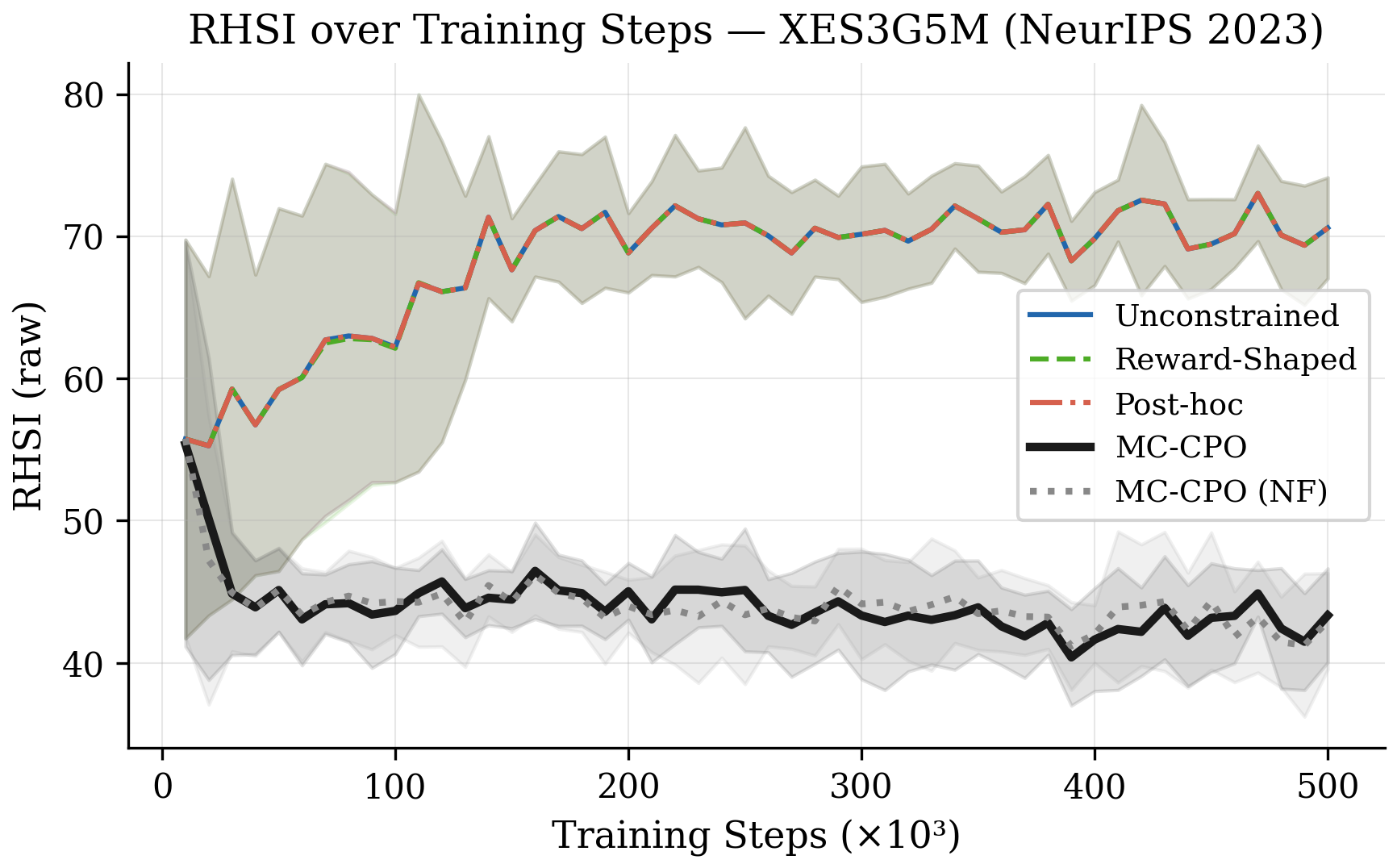}
\caption{RHSI over training steps on XES3G5M (mean $\pm$ std, 10 seeds). MC-CPO reduces RHSI by 39.8\% relative to all baselines.}
\label{fig:xes3g5m_rhsi}
\end{figure}

\subsection{Mastery Estimation Robustness}
\label{sec:noise}

To test robustness to mastery estimation error, we inject Gaussian noise $\sigma \in \{0.0, 0.05, 0.1, 0.2\}$ onto $K_t$ at inference time only. MC-CPO degrades gracefully: return drops less than $0.4\%$ at $\sigma=0.2$ and constraint satisfaction holds at 100\% across all noise levels (full results in Supplementary Appendix~C). BKT estimation error in practice falls in the range 0.05--0.15~\cite{CorbettAnderson1994BKT}, suggesting $\sigma=0.1$ is a realistic perturbation; robustness at this level provides practical assurance for deployment under estimated mastery.

\section{Discussion and Limitations}
\label{sec:discussion}

\subsection{Discussion}

Across both tabular and neural regimes, the empirical results reinforce a structural distinction between reward modification and constraint-aware optimization.

Unconstrained and reward-shaped policies achieve high engagement return but elevated RHSI, indicating systematic decoupling between engagement and mastery. In the neural regime, reward shaping is active yet converges to the unconstrained fixed point, underscoring a fundamental limitation of linear penalty methods: scalarized objectives do not guarantee satisfaction of discounted safety constraints when engagement incentives dominate.

Post-hoc filtering eliminates infeasible actions at evaluation time but does not modify the training objective. In the neural configuration, where feasibility masking is applied uniformly at evaluation, post-hoc and unconstrained policies share identical asymptotic metrics. This illustrates a structural limitation of ex post correction: filtering after optimization cannot reshape the policy landscape explored during learning, and therefore cannot close the reward--safety gap induced during training.

In contrast, MC-CPO embeds mastery-conditioned feasibility directly into the constrained optimization problem. Across environments, MC-CPO reduces RHSI while preserving competitive engagement return and maintaining discounted costs within specified tolerance thresholds. The extended chain results further demonstrate that this separation persists under multi-step interaction and stochastic mastery transitions, suggesting that the safety gap is not an artifact of minimal constructions.

The frontier mixing ablation shows that feasibility expansion events occur frequently (${\approx}0.50$ per update), though asymptotic performance differences between frontier-enabled and frontier-disabled variants are modest, suggesting frontier mixing primarily stabilizes early exploration. The 25-concept experiments confirm that MC-CPO remains effective at realistic curriculum scale while reward shaping collapses due to negligible per-step mastery signal. The mastery noise analysis (Supplementary Appendix~C) confirms robustness to estimation error up to $\sigma=0.2$. Taken together, these findings support the central thesis: mitigating reward hacking in instructional RL requires structural constraint modelling rather than additive reward penalties or ex post action masking.

\subsection{Limitations}
Several limitations qualify the scope of the present results.

\textbf{Simulation fidelity and real-world scope.}
The tabular experiments rely on stylized mastery dynamics and synthetic engagement rewards. These abstractions isolate the reward--learning alignment problem but do not capture heterogeneous learner behavior, non-stationary preferences, or long-horizon educational effects observed in real deployments. The present work addresses this limitation through validation on two real-world ITS datasets (Section~\ref{sec:real_world}): the Junyi Academy dataset (16.2M interactions from a deployed Taiwanese e-learning platform) and XES3G5M (NeurIPS 2023; 5.1M interactions from a Chinese mathematics tutoring system). Both datasets confirm that the engagement--learning decoupling pattern modelled by MC-CPO is directly observable in deployed systems. Live-learner deployment nonetheless requires IRB-approved protocols, online mastery estimation, and monitoring of feasibility restriction rates across learner subgroups steps beyond the scope of the present work.

\textbf{Mastery observability.}
The theoretical analysis assumes mastery is observed. In practice, mastery must be estimated via models such as Bayesian Knowledge Tracing or neural knowledge tracing. Estimation error may induce feasibility misclassification, potentially restricting valid actions or admitting premature progression. Although the structural masking mechanism remains well-defined under estimated mastery, performance guarantees may depend on estimator calibration.

\textbf{Function approximation gap.}
Convergence guarantees are established in the finite tabular regime. Neural experiments enforce constraint satisfaction up to a tolerance threshold to account for stochastic gradient noise and approximation error. Extending formal primal--dual guarantees to deep function approximation remains an open theoretical problem.

\textbf{Budget calibration.}
Constraint budgets in neural experiments are defined relative to unconstrained baseline costs to ensure scale invariance across synthetic environments. This yields a relative safety criterion: a policy satisfying $d_i = \kappa_i J_{c_i}(\pi_{\mathrm{unc}})$ is safer than the unconstrained baseline by construction, but 
the absolute pedagogical meaning of that safety level depends on how unsafe the unconstrained baseline itself is. In real ITS deployments, budgets would typically be grounded in domain knowledge for instance, requiring measurable mastery progress 
in at least a specified fraction of instructional windows rather than anchored to observed baseline behaviour. Translating such domain-grounded thresholds into the CMDP budget formulation constitutes an important direction for future deployment work.

\textbf{Equity considerations.}
Prerequisite-based feasibility expansion may differentially affect learners who acquire concepts at varying rates. A learner with slower mastery progression will experience a more restricted feasible action space for longer, potentially receiving less diverse instruction than a faster peer. This differential access 
constitutes an equity concern distinct from traditional algorithmic fairness, as it arises from the structure of the mastery-conditioned constraint rather than from bias in the reward signal. Future work should audit feasibility progression 
across simulated learner subgroups defined by different mastery trajectories and consider fairness-aware constraint formulations that bound differential restriction across learner profiles.

\subsection{Future Work}

Future research should extend MC-CPO to partially observable instructional settings and incorporate richer cognitive diagnostic models such as Deep Knowledge Tracing~\cite{Piech2015DKT}. The present work validates MC-CPO on two real-world ITS datasets (Junyi Academy and XES3G5M); extending to additional platforms with diverse learner populations and curriculum structures would strengthen cross-domain generalizability. Formal analysis of constrained policy optimization under deep function approximation and fairness-aware feasibility constraints represents an important theoretical direction. Deployment in live instructional systems would additionally require online mastery estimation, IRB oversight, and monitoring of feasibility restriction rates across learner subgroups.

\section{Conclusion}
\label{sec:conclusion}

This paper introduced Mastery-Conditioned Constrained Policy Optimization (MC-CPO), a structural framework for mitigating reward hacking in instructional reinforcement learning. By embedding prerequisite feasibility and discounted safety constraints directly into the CMDP, MC-CPO separates engagement maximization from pedagogical safety without relying on scalar reward penalties or ex post filtering. Theoretical analysis establishes conditions for stable primal--dual optimization, and empirical validation across minimal tabular, extended multi-step tabular, simulated neural (15 and 25 concepts), and two real-world ITS environments demonstrates improved reward--safety tradeoffs relative to unconstrained and reward-shaped baselines.

Real-world validation on the Junyi Academy dataset (16.2M interactions) and the XES3G5M benchmark (NeurIPS 2023, 5.1M interactions) confirms that the engagement--learning decoupling targeted by MC-CPO is directly observable in deployed systems, at rates of 26.5\% and 3.1\% of student interactions respectively. MC-CPO reduces RHSI by 18.4\% and 39.8\% on the two datasets, increases mean per-episode mastery gain $\Delta K$ by 18.3\% and 54.0\%, and activates non-zero Lagrange multipliers on both confirming structural constraint enforcement under three independent measurement frameworks. These results support the central thesis of this work: modeling pedagogical feasibility as a structural constraint, rather than as a reward modification, provides a principled and empirically validated foundation for safer adaptive instructional policies in real deployed intelligent tutoring systems.


\section*{Declaration of Competing Interests}
The authors declare that they have no known competing financial interests
or personal relationships that could have appeared to influence the work
reported in this paper.

\section*{Data Availability}
The two real-world datasets used in this study are publicly available.
The Junyi Academy Learning Activity dataset is available at
\url{https://www.kaggle.com/datasets/junyiacademy/learning-activity-public-dataset-by-junyi-academy}.
The XES3G5M dataset is available via the NeurIPS 2023 benchmark
repository \cite{Liu2023XES3G5M}. Simulated environment code
will be made available upon acceptance.

\section*{Ethics Statement}
All experiments in this paper are conducted using publicly released,
anonymised interaction logs from deployed intelligent tutoring systems.
No new data collection involving human participants was undertaken.
The Junyi Academy dataset and the XES3G5M dataset are available for
academic research purposes without restriction.
IRB approval is not required for analysis of pre-existing, publicly
released, and fully anonymised educational datasets.



\bibliographystyle{elsarticle-num}
\bibliography{references}

\begin{thebibliography}{10}
\expandafter\ifx\csname url\endcsname\relax
  \def\url#1{\texttt{#1}}\fi
\expandafter\ifx\csname urlprefix\endcsname\relax\def\urlprefix{URL }\fi
\expandafter\ifx\csname href\endcsname\relax
  \def\href#1#2{#2} \def\path#1{#1}\fi

\bibitem{VanLehn2011ITS}
K.~VanLehn, The relative effectiveness of human tutoring, intelligent tutoring systems, and other tutoring systems, Educational Psychologist 46~(4) (2011) 197--221.
\newblock \href {https://doi.org/10.1080/00461520.2011.611369} {\path{doi:10.1080/00461520.2011.611369}}.

\bibitem{Koedinger2006CognitiveTutors}
K.~R. Koedinger, A.~T. Corbett, Cognitive tutors: Technology bringing learning sciences to the classroom, in: R.~K. Sawyer (Ed.), The {Cambridge} Handbook of the Learning Sciences, Cambridge University Press, 2006, pp. 61--77.

\bibitem{SuttonBarto2018}
R.~S. Sutton, A.~G. Barto, \href{http://incompleteideas.net/book/the-book-2nd.html}{Reinforcement Learning: An Introduction}, 2nd Edition, MIT Press, Cambridge, MA, 2018.
\newline\urlprefix\url{http://incompleteideas.net/book/the-book-2nd.html}

\bibitem{Doroudi2019WheresReward}
S.~Doroudi, V.~Aleven, E.~Brunskill, Where's the reward? {A} review of reinforcement learning for instructional sequencing, International Journal of Artificial Intelligence in Education 29~(4) (2019) 568--620.
\newblock \href {https://doi.org/10.1007/s40593-019-00187-x} {\path{doi:10.1007/s40593-019-00187-x}}.

\bibitem{Mon2023RLInEducation}
B.~F. Mon, A.~Wasfi, M.~Hayajneh, A.~Slim, N.~A. Ali, Reinforcement learning in education: A literature review, Informatics 10~(3) (2023) 74.
\newblock \href {https://doi.org/10.3390/informatics10030074} {\path{doi:10.3390/informatics10030074}}.

\bibitem{Liu2023XES3G5M}
Z.~Liu, Q.~Liu, T.~Guo, J.~Chen, J.~Zhu, J.~Tang, W.~Luo, {XES3G5M}: A knowledge tracing benchmark dataset with auxiliary information, in: Advances in Neural Information Processing Systems (NeurIPS), 2023.

\bibitem{Amodei2016ConcreteProblems}
D.~Amodei, C.~Olah, J.~Steinhardt, P.~Christiano, J.~Schulman, D.~Man{\'e}, \href{https://arxiv.org/abs/1606.06565}{Concrete problems in {AI} safety}, arXiv preprint arXiv:1606.06565 (2016).
\newline\urlprefix\url{https://arxiv.org/abs/1606.06565}

\bibitem{Skalse2022RewardHacking}
J.~Skalse, N.~H.~R. Howe, D.~Krasheninnikov, D.~Krueger, \href{https://arxiv.org/abs/2209.13085}{Defining and characterizing reward hacking}, arXiv preprint arXiv:2209.13085 (2022).
\newblock \href {https://doi.org/10.48550/arXiv.2209.13085} {\path{doi:10.48550/arXiv.2209.13085}}.
\newline\urlprefix\url{https://arxiv.org/abs/2209.13085}

\bibitem{HadfieldMenell2017IRD}
D.~Hadfield-Menell, S.~Milli, P.~Abbeel, S.~J. Russell, A.~Dragan, \href{https://proceedings.neurips.cc/paper/2017/hash/32fdab6559cdfa4f167f8c31b9199643-Abstract.html}{Inverse reward design}, in: Advances in Neural Information Processing Systems ({NeurIPS}), Vol.~30, 2017.
\newline\urlprefix\url{https://proceedings.neurips.cc/paper/2017/hash/32fdab6559cdfa4f167f8c31b9199643-Abstract.html}

\bibitem{Shihab2025ProxyGaming}
I.~F. Shihab, S.~Akter, A.~Sharma, \href{https://arxiv.org/abs/2507.05619}{Detecting proxy gaming in {RL} and {LLM} alignment via evaluator stress tests}, arXiv preprint arXiv:2507.05619 (2025).
\newline\urlprefix\url{https://arxiv.org/abs/2507.05619}

\bibitem{Altman1999CMDP}
E.~Altman, Constrained {Markov} Decision Processes, Chapman and Hall/{CRC}, 1999.

\bibitem{Puterman1994MDP}
M.~L. Puterman, Markov Decision Processes: Discrete Stochastic Dynamic Programming, John Wiley \& Sons, 1994.
\newblock \href {https://doi.org/10.1002/9780470316887} {\path{doi:10.1002/9780470316887}}.

\bibitem{Achiam2017CPO}
J.~Achiam, D.~Held, A.~Tamar, P.~Abbeel, \href{https://proceedings.mlr.press/v70/achiam17a.html}{Constrained policy optimization}, in: Proceedings of the 34th International Conference on Machine Learning ({ICML}), Vol.~70 of Proceedings of Machine Learning Research, {PMLR}, 2017, pp. 22--31.
\newline\urlprefix\url{https://proceedings.mlr.press/v70/achiam17a.html}

\bibitem{Chow2018Lyapunov}
Y.~Chow, O.~Nachum, E.~Du{\'e}nez-Guzm{\'a}n, M.~Ghavamzadeh, \href{https://proceedings.neurips.cc/paper/2018/hash/4fe5149039b52765bde64beb9f674940-Abstract.html}{A {Lyapunov}-based approach to safe reinforcement learning}, in: Advances in Neural Information Processing Systems ({NeurIPS}), Vol.~31, 2018.
\newline\urlprefix\url{https://proceedings.neurips.cc/paper/2018/hash/4fe5149039b52765bde64beb9f674940-Abstract.html}

\bibitem{Tessler2019RCPO}
C.~Tessler, D.~J. Mankowitz, S.~Mannor, \href{https://arxiv.org/abs/1805.11074}{Reward constrained policy optimization}, in: International Conference on Learning Representations ({ICLR}), 2019.
\newline\urlprefix\url{https://arxiv.org/abs/1805.11074}

\bibitem{Wachi2024SafeRLSurvey}
A.~Wachi, X.~Shen, Y.~Sui, \href{https://www.ijcai.org/proceedings/2024/0913.pdf}{A survey of constraint formulations in safe reinforcement learning}, in: Proceedings of the Thirty-Third International Joint Conference on Artificial Intelligence ({IJCAI-24}), 2024.
\newline\urlprefix\url{https://www.ijcai.org/proceedings/2024/0913.pdf}

\bibitem{Zhao2023StatewiseSafeRL}
W.~Zhao, T.~He, R.~Chen, T.~Wei, C.~Liu, \href{https://arxiv.org/abs/2302.03122}{State-wise safe reinforcement learning: A survey}, arXiv preprint arXiv:2302.03122 (2023).
\newline\urlprefix\url{https://arxiv.org/abs/2302.03122}

\bibitem{Yang2023FPI}
Y.~Yang, Z.~Zheng, S.~E. Li, J.~Duan, J.~Liu, X.~Zhan, Y.-Q. Zhang, \href{https://arxiv.org/abs/2304.08845}{Feasible policy iteration for safe reinforcement learning}, arXiv preprint arXiv:2304.08845 (2023).
\newline\urlprefix\url{https://arxiv.org/abs/2304.08845}

\bibitem{CorbettAnderson1994BKT}
A.~T. Corbett, J.~R. Anderson, Knowledge tracing: Modeling the acquisition of procedural knowledge, User Modeling and User-Adapted Interaction 4~(4) (1994) 253--278.
\newblock \href {https://doi.org/10.1007/BF01099821} {\path{doi:10.1007/BF01099821}}.

\bibitem{Piech2015DKT}
C.~Piech, J.~Bassen, J.~Huang, S.~Ganguli, M.~Sahami, L.~Guibas, J.~Sohl-Dickstein, \href{https://proceedings.neurips.cc/paper/2015/hash/bac9162b47c56fc8a4d2a519803d51b3-Abstract.html}{Deep knowledge tracing}, in: Advances in Neural Information Processing Systems ({NeurIPS}), Vol.~28, 2015.
\newline\urlprefix\url{https://proceedings.neurips.cc/paper/2015/hash/bac9162b47c56fc8a4d2a519803d51b3-Abstract.html}

\bibitem{Deng2023GITS}
Y.~Deng, J.~Ren, \href{https://arxiv.org/abs/2312.10053}{Towards goal-oriented intelligent tutoring systems in online education}, arXiv preprint arXiv:2312.10053 (2023).
\newline\urlprefix\url{https://arxiv.org/abs/2312.10053}

\bibitem{Kushwaha2025Survey}
A.~Kushwaha, K.~Ravish, P.~Lamba, P.~Kumar, \href{https://arxiv.org/abs/2505.17342}{A survey of safe reinforcement learning and constrained {MDP}s: State of the art and open problems}, arXiv preprint arXiv:2505.17342 (2025).
\newline\urlprefix\url{https://arxiv.org/abs/2505.17342}

\bibitem{Garcia2015SafeRLSurvey}
J.~Garc{\'i}a, F.~Fern{\'a}ndez, \href{https://www.jmlr.org/papers/v16/garcia15a.html}{A comprehensive survey on safe reinforcement learning}, Journal of Machine Learning Research 16~(42) (2015) 1437--1480.
\newline\urlprefix\url{https://www.jmlr.org/papers/v16/garcia15a.html}

\bibitem{Gu2023SafeRLReview}
S.~Gu, L.~Yang, Y.~Du, G.~Chen, F.~Walter, J.~Wang, A.~Knoll, A review of safe reinforcement learning: Methods, theories, and applications, {IEEE} Transactions on Pattern Analysis and Machine Intelligence 46~(5) (2024) 3489--3508.
\newblock \href {https://doi.org/10.1109/TPAMI.2023.3272206} {\path{doi:10.1109/TPAMI.2023.3272206}}.

\bibitem{Dalal2018SafeExploration}
G.~Dalal, K.~Dvijotham, M.~Vecerik, T.~Hester, C.~Paduraru, Y.~Tassa, \href{https://openreview.net/forum?id=SkfrvsA9YX}{Safe exploration in continuous action spaces}, in: International Conference on Learning Representations ({ICLR}), 2018.
\newline\urlprefix\url{https://openreview.net/forum?id=SkfrvsA9YX}

\bibitem{Wachi2023SafeExploration}
A.~Wachi, W.~Hashimoto, X.~Shen, K.~Hashimoto, \href{https://openreview.net/forum?id=dQLsvKNwZC}{Safe exploration in reinforcement learning: A generalized formulation}, in: Advances in Neural Information Processing Systems ({NeurIPS}), 2023.
\newline\urlprefix\url{https://openreview.net/forum?id=dQLsvKNwZC}

\bibitem{Yao2023CCPO}
Y.~Yao, H.~Liu, G.~Zhou, L.~Fan, Y.~Zhu, \href{https://arxiv.org/abs/2310.03718}{Constraint-conditioned policy optimization for versatile safe reinforcement learning}, in: Advances in Neural Information Processing Systems ({NeurIPS}), Vol.~36, 2023.
\newline\urlprefix\url{https://arxiv.org/abs/2310.03718}

\bibitem{Bura2021DOPE}
A.~Bura, A.~HasanzadeZonuzy, D.~Kalathil, S.~Shakkottai, J.-F. Chamberland, \href{https://arxiv.org/abs/2112.00885}{{DOPE}: Doubly optimistic and pessimistic exploration for safe reinforcement learning}, arXiv preprint arXiv:2112.00885 (2021).
\newline\urlprefix\url{https://arxiv.org/abs/2112.00885}

\bibitem{Schulman2017PPO}
J.~Schulman, F.~Wolski, P.~Dhariwal, A.~Radford, O.~Klimov, \href{https://arxiv.org/abs/1707.06347}{Proximal policy optimization algorithms}, Technical report, OpenAI (2017).
\newline\urlprefix\url{https://arxiv.org/abs/1707.06347}

\bibitem{Pan2022RewardMisspec}
A.~Pan, K.~Bhatia, J.~Steinhardt, \href{https://openreview.net/forum?id=JYtwGwIL7ye}{The effects of reward misspecification: Mapping and mitigating misaligned models}, in: International Conference on Learning Representations ({ICLR}), 2022.
\newline\urlprefix\url{https://openreview.net/forum?id=JYtwGwIL7ye}

\bibitem{Gao2023ScalingReward}
L.~Gao, J.~Schulman, J.~Hilton, \href{https://proceedings.mlr.press/v202/gao23h.html}{Scaling laws for reward model overoptimization}, in: Proceedings of the 40th International Conference on Machine Learning ({ICML}), Vol. 202 of Proceedings of Machine Learning Research, {PMLR}, 2023, pp. 10835--10866.
\newline\urlprefix\url{https://proceedings.mlr.press/v202/gao23h.html}

\bibitem{Opryshko2025MCVL}
E.~Opryshko, U.~Jain, I.~Gilitschenski, \href{https://openreview.net/forum?id=OmYqzp8NO7}{Modification-considering value learning for reward hacking mitigation in {RL}}, arXiv preprint (2025).
\newline\urlprefix\url{https://openreview.net/forum?id=OmYqzp8NO7}

\bibitem{Ng1999RewardShaping}
A.~Y. Ng, D.~Harada, S.~Russell, Policy invariance under reward transformations: Theory and application to reward shaping, in: Proceedings of the 16th International Conference on Machine Learning ({ICML}), 1999, pp. 278--287.

\bibitem{Christiano2017RLHF}
P.~Christiano, J.~Leike, T.~B. Brown, M.~Martic, S.~Legg, D.~Amodei, \href{https://proceedings.neurips.cc/paper/2017/hash/d5e2c0adad503c91f91df240d0cd4e49-Abstract.html}{Deep reinforcement learning from human preferences}, in: Advances in Neural Information Processing Systems ({NeurIPS}), Vol.~30, 2017.
\newline\urlprefix\url{https://proceedings.neurips.cc/paper/2017/hash/d5e2c0adad503c91f91df240d0cd4e49-Abstract.html}

\bibitem{Ibarz2018RewardModelHacking}
B.~Ibarz, J.~Leike, T.~Pohlen, G.~Irving, S.~Legg, D.~Amodei, \href{https://proceedings.neurips.cc/paper/2018/hash/8cbe9ce23f42628c98f80fa0fac8b19a-Abstract.html}{Reward learning from human preferences and demonstrations in {Atari}}, in: Advances in Neural Information Processing Systems ({NeurIPS}), Vol.~31, 2018.
\newline\urlprefix\url{https://proceedings.neurips.cc/paper/2018/hash/8cbe9ce23f42628c98f80fa0fac8b19a-Abstract.html}

\bibitem{Dai2024SafeRLHF}
J.~Ji, M.~Liu, J.~Dai, X.~Pan, C.~Zhang, C.~Bian, C.~Zhang, R.~Sun, Y.~Wang, Y.~Yang, \href{https://openreview.net/forum?id=TyFrPOKYXw}{Safe {RLHF}: Safe reinforcement learning from human feedback}, in: International Conference on Learning Representations ({ICLR}), 2024.
\newline\urlprefix\url{https://openreview.net/forum?id=TyFrPOKYXw}

\bibitem{Rafferty2011POMDP}
A.~N. Rafferty, E.~Brunskill, T.~L. Griffiths, P.~Shafto, Faster teaching via {POMDP} planning, in: Artificial Intelligence in Education, Springer, 2011, pp. 280--287.

\bibitem{Ammanabrolu2020KGA2C}
P.~Ammanabrolu, M.~Hausknecht, \href{https://openreview.net/forum?id=B1x6w0EtwH}{Graph constrained reinforcement learning for natural language action spaces}, in: International Conference on Learning Representations ({ICLR}), 2020.
\newline\urlprefix\url{https://openreview.net/forum?id=B1x6w0EtwH}

\bibitem{Borkar2008SA}
V.~S. Borkar, Stochastic Approximation: A Dynamical Systems Viewpoint, Cambridge University Press, 2008.

\bibitem{JunyiAcademy2015}
{Junyi Academy}, \href{https://www.kaggle.com/datasets/junyiacademy/ learning-activity-public-dataset-by-junyi-academy}{Learning activity public dataset} (2015).
\newline\urlprefix\url{https://www.kaggle.com/datasets/junyiacademy/ learning-activity-public-dataset-by-junyi-academy}

\end{thebibliography}

\noindent\textit{Supplementary Material.} Full proofs, neural training curves, mastery estimation robustness results, and the frontier-conditioned exploration derivation are provided in the supplementary file submitted alongside this manuscript.

\end{document}